\documentclass[runningheads]{llncs}
\usepackage{graphicx}
\usepackage{bbm}

\usepackage{tikz}
\usepackage{comment}
\usepackage{amsmath,amssymb} %

\usepackage{color}
\usepackage{multirow}
\usepackage{booktabs} %
\usepackage[percent]{overpic}

\usepackage[accsupp]{axessibility} %

\usepackage{sidecap} %
\sidecaptionvpos{figure}{t} %

\usepackage{breakcites} %
\usepackage{hyperref}
\hypersetup{
    colorlinks=true,
    linkcolor=blue,
    filecolor=magenta,      
    urlcolor=cyan,
    pdftitle={Overleaf Example},
    pdfpagemode=FullScreen,
    }
\newcommand{\tildeapprox}{{\raise.17ex\hbox{$\scriptstyle\sim$}}}
\newcommand{\secref}[1]{Sec. \ref{#1}}
\newcommand{\figref}[1]{Fig. \ref{#1}}

\renewcommand{\eqref}[1]{Eq.~(\ref{#1})}

\definecolor{atomictangerine}{rgb}{0.8, 0.2, 0.1}
\definecolor{turq}{rgb}{0.0, 0.5, 0.5}
\definecolor{darkturq}{rgb}{0.0, 0.4, 0.4}
\definecolor{bright}{rgb}{0.8, 0.1, 0}
\definecolor{darkgray}{gray}{0.3}
\definecolor{gray}{gray}{0.5}
\definecolor{mahogany}{rgb}{0.6, 0.05, 0.05}
\definecolor{myblue}{rgb}{0.3,0.05,0.9}
\definecolor{darkgreen}{rgb}{0.1,0.5,0.0}

\definecolor{olive}{rgb}{0.537, 0.627, 0.318}
\definecolor{green}{rgb}{0.22, 0.463, 0.114}
\definecolor{grey}{rgb}{0.4, 0.4, 0.4}
\definecolor{blue}{rgb}{0.135, 0.1, 0.663}
\definecolor{pink}{rgb}{0.761, 0.482, 0.627}
\definecolor{brightpink}{rgb}{0.8, 0.0, 0.4}
\definecolor{darkpink}{rgb}{0.561, 0.282, 0.427}
\definecolor{black}{rgb}{0., 0., 0.}

\newcommand\ignore[1]{}
\newcommand\gal[1]{}
\newcommand\yuval[1]{}
\newcommand\niv[1]{}
\newcommand\eli[1]{}
\newcommand\rinon[1]{}
\newcommand\edit[1]{#1}

\newcommand{\cossim}[2]{\text{cos}(#1, #2)}

\renewcommand\vec[1]{\mathbf{#1}}

\newcommand{\ftheta}{f_\theta}

\newcommand{\A}{\vec{A}}
\newcommand{\w}{\vec{w}}

\newcommand{\VL}{V\&L {}}

\newcommand{\W}{\mathcal{W}}
\newcommand{\I}{\mathcal{I}}
\newcommand{\T}{\mathcal{T}}
\newcommand{\CLIPI}{h^{\mathcal{I}}}
\newcommand{\CLIPT}{h^{\mathcal{T}}}

\newcommand{\PerVL}{PerVL {}}
\newcommand{\img}{I}
\newcommand{\concept}{[CONCEPT]}

\newcommand{\etal}{{{et al}. }}

\begin{document}
\pagestyle{headings}
\mainmatter
\def\ECCVSubNumber{8098}  %

\title{``This is my unicorn, Fluffy'': Personalizing frozen vision-language representations}

\titlerunning{``This is my unicorn, Fluffy'':  Personalizing frozen V\&L representations}
\author{Niv Cohen \inst{1,2} \and
Rinon Gal\inst{1,3} \and
Eli A. Meirom\inst{1} \and \\
Gal Chechik\inst{1,4} \and
Yuval Atzmon\inst{1}}

\authorrunning{N. Cohen et al.}
\institute{NVIDIA Research, Israel \and
The Hebrew University of Jerusalem, Jerusalem, Israel \and
Tel Aviv University, Israel \and Bar Ilan University, Israel}

\maketitle

\begin{abstract}
Large Vision \& Language models pretrained on web-scale data provide representations that are invaluable for numerous V\&L problems. However, it is unclear how they can be extended to reason about \textit{user-specific} visual concepts in unstructured language. This problem arises in multiple domains, from personalized image retrieval to personalized interaction with smart devices.
We introduce a new learning setup called {\em Personalized Vision \& Language} (PerVL) with two new benchmark datasets for retrieving and segmenting user-specific (``personalized'') concepts ``in the wild''. In PerVL, one should learn personalized concepts (1) {\em independently}  of the downstream task (2) allowing a pretrained model to reason about them with free language, and (3) without providing personalized negative examples.
We propose an architecture for solving PerVL that operates by \textit{expanding} the input vocabulary of a pretrained model with new word embeddings for the  personalized concepts. The model can then simply employ them as part of a sentence.
We demonstrate that our approach learns personalized visual concepts from a few examples and effectively applies them in image retrieval and semantic segmentation using rich textual queries. For example the model improves MRR by 51.1\% (28.4\% vs 18.8\%) compared to the strongest baseline.

The code and benchmark are available on github under \href{https://github.com/NVlabs/PALAVRA}{NVlabs/PALAVRA} and \href{https://github.com/NVlabs/PerVLBenchmark}{NVlabs/PerVLBenchmark}.

\end{abstract}

\section{Introduction}
Large  Vision \& Language (\VL\!\!) models pre-trained on web-scale data made a breakthrough in computer vision \cite{radford2021learning,yuan2021florence,bommasani2021opportunities}. These models  provide a multimodal vision-language representation, and are used in a multitude of downstream tasks, from image captioning \cite{mokady2021clipcap} and video retrieval \cite{fang2021clip2video}, through image generation \cite{gal2021stylegan,patashnik2021styleclip} and segmentation~\cite{zabari2021semantic,li2022languagedriven}, to robotic manipulation~\cite{shridhar2021cliport}. All these tasks benefit from the ``open-world'' capabilities of large V\&L models, enabling the use of rich, free-form text with a long ``tail" vocabulary of visual categories.

However, even with these powerful representations, an important challenge remains: How can these models be leveraged to reason about \textit{user-specific}, ``\textit{personalized}'' object instances in open-world vision problems? For example, we may wish to find an image that portrays us wearing a \textit{specific} sweater, ask a robot assistant to make us coffee in \textit{our} ``best-mom mug", or synthesize an image of our child's treasured toy Fluffy in an entirely new context.

\begin{figure}[t]
    \centering
    \includegraphics[width=0.99\textwidth, trim={0.2cm 0cm 0cm 0cm},clip]{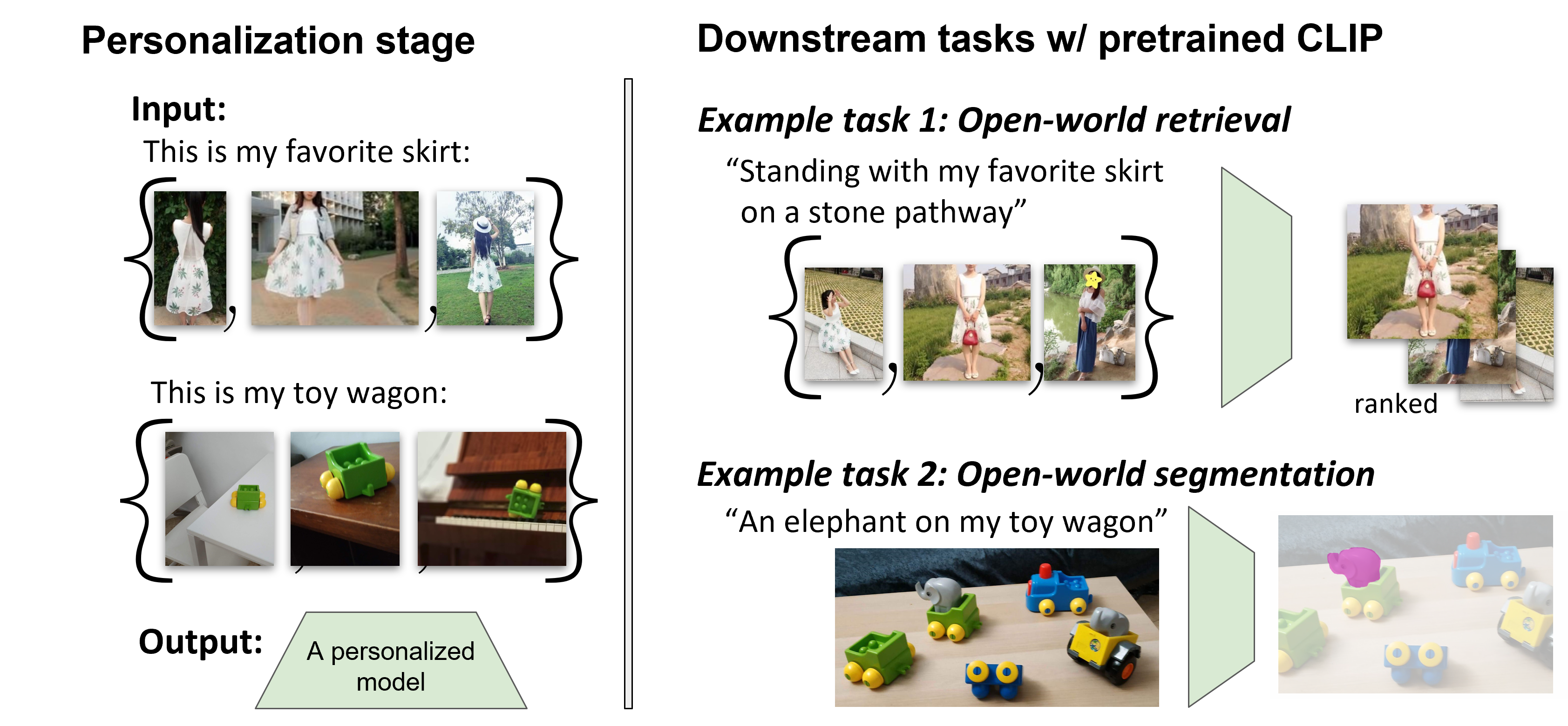} %
    \caption{
    \textbf{The Personalized Vision \& Language (\PerVL\!\!) learning setup}. \textbf{Left:} A user provides a few image examples of their personalized visual concepts: a favorite skirt (top), or a toddler's toy wagon (bottom). Examples are provided independently of the downstream tasks. \textbf{(Right)} the personalized model can be used in various downstream tasks. \textbf{Top-right:} An image retrieval task: given a textual query and a collection of images,
    rank and retrieve the queried image. \textbf{(Bottom-right)} Open-world semantic segmentation task. Segment a personalized object referred by a textual query. %
    This example illustrates multiple ambiguities. First, there are two wagons that carry an elephant. Second, there are two wagons that correspond to the personalized concept. Resolving the ambiguity requires reasoning in both the visual and text modalities.
    }
    \label{fig1}
\end{figure}

Clearly, pretrained \VL models cannot be used directly to reason about new personal items. 
Luckily, it is typically easy for a user to collect a few image examples for a personalized concept. %
It then remains to develop methods that extend the pretrained models to new concepts using these examples.
One challenge %
is that
typically it is easy for people to provide positive image examples for a concept, but harder to provide consistent negative distractor examples \cite{hsieh2019classification,shinoda2020binary}.

Learning from a few examples is considered a hallmark of intelligence. When people learn novel concepts from a few examples \cite{malaviya2022can,carey1978acquiring,lake2020people,markman1990constraints,markman2003use}, they can seamlessly employ them in their semantic mental state and \textit{reason jointly} both over the personalized concepts and over a large body of prior knowledge. Could a computational approach learn in a similar way using a pretrained \VL model?

Previous efforts  \cite{skantze2021collie,zhang2021tip,gao2021clip,zhou2021learning} focused on learning a transformation module on top of CLIP's output space. However, as we explain below, these approaches risk forgetting prior knowledge, or face difficulties in accessing it concurrently with newly learned concepts.
In addition, these previous approaches take a multiclass approach, discriminating between several new concepts. They are not designed for learning a single new personalized concept, which is natural in the context of personalization.
Therefore, it is unknown how to learn a single personalized concept from few image examples in a way that (1) allows the pretrained model to reason about the concept with free language, and (2) uses only ``positive" image examples of the target concept.

Here, we address the question of personalizeing a pretrained model using few samples while maintaining its performance on the original vocabulary.
We study a new representation learning setup, which we call ``\textit{Personalized Vision \& Language}'' (\PerVL\!\!)  (\figref{fig1}). In \PerVL\!\!, we are given a pretrained \VL model, one or more personalized visual concepts, a few training images of each concept, and a string describing the concept type, like ``a mug'' or ``a short sleeve top''.
The goal is to learn a representation that can later be used to solve a set of  downstream \VL tasks involving the personalized concept. No further supervision is given for these downstream tasks. \PerVL arises in various scenarios. In image retrieval, a user may tag a few of their images and wish to retrieve other photos of that concept in a visual specific context \cite{chen2020image,anwaar2021compositional}; in human-robot interaction, a worker may show a specific tool to a robotic arm, and instruct how to  use it \cite{shridhar2021cliport,wang2022hierarchical,lynch2020language}; in video applications, an operator may search for a specific known item in the context of other items or people doing activities that are described with language.

Unlike previous efforts, instead of modifying a \VL model \textit{output}, we propose a framework for expanding its \textit{input} vocabulary.  Specifically, we learn new word embeddings for the new personalized concepts by intervening with the model input space. The concept of ``my best-mom mug'' would be associated with a new symbol [MY BEST-MOM MUG] that has its own \textit{dense} word embedding. The model could later represent sentences that use it, like ``\textit{Sipping tea from my best-mom mug on a porch}" by detecting ``my best mom mug" and mapping its symbol [MY BEST-MOM MUG] to its new embedding vector. Such tokens trivially preserve the structure of the original model, since the encoder model itself remains unmodified. Moreover, as we show below, the new concepts can also be easily integrated into existing downstream \VL tasks. In summary, we address the question of using a small number of samples to personalize a pretrained \VL model, while maintaining its performance in the original vocabulary.

This paper makes the following novel contributions: (1) A new
representation learning setup, \textit{\PerVL}\!\!, for personalizing  \VL representations, while keeping their ``zero-shot'' reasoning capabilities. (2) Two new benchmark datasets for \textit{\PerVL\!\!}.
(3) A novel approach, PALAVRA\footnote[7]{Palavra means ``word'' in Portuguese, \textit{as we learn new word-embeddings.} %
For acronym lovers, PALAVRA also stands for ``Personalizing LAnguage Vision RepresentAtions''},
to \textit{expand} and personalize the vocabulary of the \VL representation \textit{inputs}. PALAVRA uses a cycle-consistent loss,  learned with  positive image examples only.
(4) A technique for using a \textit{textual} encoder to improve the generalization of a network to new \textit{visual} concepts.

\section{Related work}
The success of CLIP led to diverse work that leverage its powerful representation for few-shot learning. Most works \cite{skantze2021collie,zhang2021tip,gao2021clip,ma2021simple} are based on learning a residual ``adapter'' layer \cite{houlsby2019parameter} over the output of CLIP encoders. Taking a different approach, \cite{zhou2021learning} proposes learning a soft prefix to improve accuracy in a classification task. Our work differs from these approaches in two key aspects:
(1) They focus solely on classifying images using a narrow vocabulary.
In contrast, our setup learns a representation which is then used in any downstream tasks. Moreover, our method \textit{expands} CLIP's vocabulary rather than narrowing it.
(2) Adapter-based methods override the output representation of the encoders, leading to a change in their input $\rightarrow$ output mappings. Our method does not change the pretrained mapping but enriches its input vocabulary with new concepts.

Recently, \cite{wortsman2021robust,kumar2022fine,hewitt2021ensembles} have shown that fine-tuning can actively harm out-of-distribution generalization, even when tested on the same downstream task for which the model was tuned. 
Our method does not fine-tune the pretrained model and does not leverage in-distribution labeled examples for the downstream tasks. %

Other approaches \cite{tsimpoukelli2021multimodal,hill2020grounded} study ``fast'' concept learning combined with ``slow-learned'' concepts, %
showing that the new concepts can be applied to ``slowly-learned'' downstream tasks. However, the ``fast'' learned concepts are stored implicitly in the network activations, rather than grounded in the  vocabulary. 

A related set of works can be found in the image captioning and generator inversion tasks. While their goals are different, these works nonetheless aim to extract meaningful semantic information from images and map them to tokens that represent the concepts - in this case, words or latent codes. Of these, a series of works \cite{chunseong2017attend,denton2015user,feng2017personalized,del2020ratt,jia2020personalized,long2020cross,shuster2019engaging} focus on personalizing image captions according to a user writing style. Alternatively,  \cite{hendricks2016deep,venugopalan2017captioning,wu2018decoupled,zheng2019intention,lu2018neural,demirel2019image} extend image captions with novel concepts using ``slot filling'', which are placeholders for nouns that are filled using object detector predictions. In inversion, models typically aim to identify codes in the latent spaces of pre-trained generators, which represent specific images or identities~\cite{abdal2019image2stylegan,richardson2021encoding}. These can then be used in downstream tasks such as editing~\cite{shen2020interfacegan,abdal2021styleflow} or super resolution~\cite{menon2020pulse}. In some cases, the model is further fine-tuned to better represent a specific instance~\cite{roich2021pivotal,alaluf2022hyperstyle,dinh2022hyperinverter,tzaban2022stitch}, or to allow more faithful replication of personalized traits such as expressions~\cite{nitzan2022mystyle}.

Our model differs  from zero and few-shot learning (FSL) based on meta-learning ~\cite{finn2017model,vinyals2016matching,sung2018learning,snell2017prototypical,chen2021meta,DAP,ALE,xianCVPR,LAGO,COSMO,ZEST} or incremental learning    \cite{tao2020few,fan2021flar,ren2020wandering,cheraghian2021semantic,khan2021personalizing,wu2021striking} in three aspects. First, we impose stronger generalization requirements. Our model can reason about new concepts in diverse downstream tasks, which may be unknown at training time. %
Second, in common FSL, the concept distribution used during meta-learning (``support set'') is also used during the FSL stage. For example, meta-learn on birds, then do FSL with new types of birds. While our technique for training with text allows to generalize beyond the domain of concepts in the training images.
Third, our approach improves upon CLIP's zero-shot perceptual capabilities, and is compatible with many CLIP-based downstream tasks. 

\edit{Finally, in existing FSL benchmarks \cite{vinyals2016matching,ren2018metalearning,zhang2021personalized} there is \textit{no} instance level annotations, and there is only a single task. As a result, existing FSL benchmarks do not directly fit our setting. Our work addresses \textit{rich text} query \textit{of a specific instance}, that can be used in a flexible way with many downstream tasks. 
 }

\section{A new setup, \textit{Personalized Vision \& Language}}
We propose ``\textit{Personalized Vision \& Language}'' (\PerVL\!\!), a new %
representation learning setup, to personalize a pretrained model with few positive image examples, without supervision for the downstream task. %

In \PerVL, we are given a pretrained model $h(S_V, \img)$ that accepts a sentence $S$ and an image $\img$. The sentences that the model accepts are defined in a vocabulary $V$. We wish to update $h$ so that it can accept sentences from an expanded vocabulary $V' = V \cup C $ where $C$ is a new set of concepts $C = \{c_1,....c_k\}$, which results in an extended model $h'(S_{V'},I)$.
In general, we expect that adapting the model would not strongly affect the original vocabulary, namely $h'(S_V,I) \approx h(S_V,I)$. %

At training (personalization) time, we adapt the model given a small set of images $\{\img_i\}_{i=1}^{N_c}$ for every concept $c$, without assuming access to negative training \textit{images}. %
We are also provided with a string describing the type of the new concept, such as a ``mug'' or a ``short sleeve top''.
Stating the type is a natural way for non-expert users to provide prior knowledge about the personalized concept. The type can be used to guide learning to distinguish the personalized concept from the general concept type.
Concepts describing coarser classes from a hierarchy of concepts (e.g. ``dog'' for ``poodle'') have been used for this purpose \cite{dekel2004large}.
We denote the concept type by $S_c$.

During inference, we are given a downstream \VL task $T$ that can be inferred using the pretrained model $h$ for the vocabulary $V$, and we wish to solve it for an instance $x$ that contains the new concept $c$. The instance may contain images and sentences pertaining to $c$.

\textbf{Encoder \PerVL}:
Here we focus on the special case of CLIP \cite{radford2021learning}. The model $h$ applies a cosine similarity between a sentence $S$ and an image $\img$: $h(S, \img) = \cossim{\CLIPT(S)}{ \CLIPI(\img)}$, where $\CLIPI$ and $\CLIPT$ are CLIP image and text encoders.

\section{Methods}
\label{sec:methods}

Before describing our approach, we first explain the reasons for expanding the \textit{input} vocabulary of the \VL model and how it differs from previous approaches.

Several studies extend CLIP by learning an ``Adapter'' module on top of the CLIP representation \cite{gao2021clip,skantze2021collie,zhang2021tip,ma2021simple,houlsby2019parameter}. That module is applied to the \textit{output} of a CLIP encoder network that is kept frozen. It is trained for a classification task with labeled data and a templated text query (``a photo of a [concept-type]'').

We show below (Sec. \ref{sec_experiments} \& Appendix \ref{sec_additional_results}) that this approach tends to be brittle and fails when its input sentences deviate from the template used for training. This is probably because the adapter \textit{overrides} the output representation of the encoder, so training it with very few examples hurts its generalization power.

\begin{SCfigure}
    \centering   
    \includegraphics[width=0.7\textwidth]{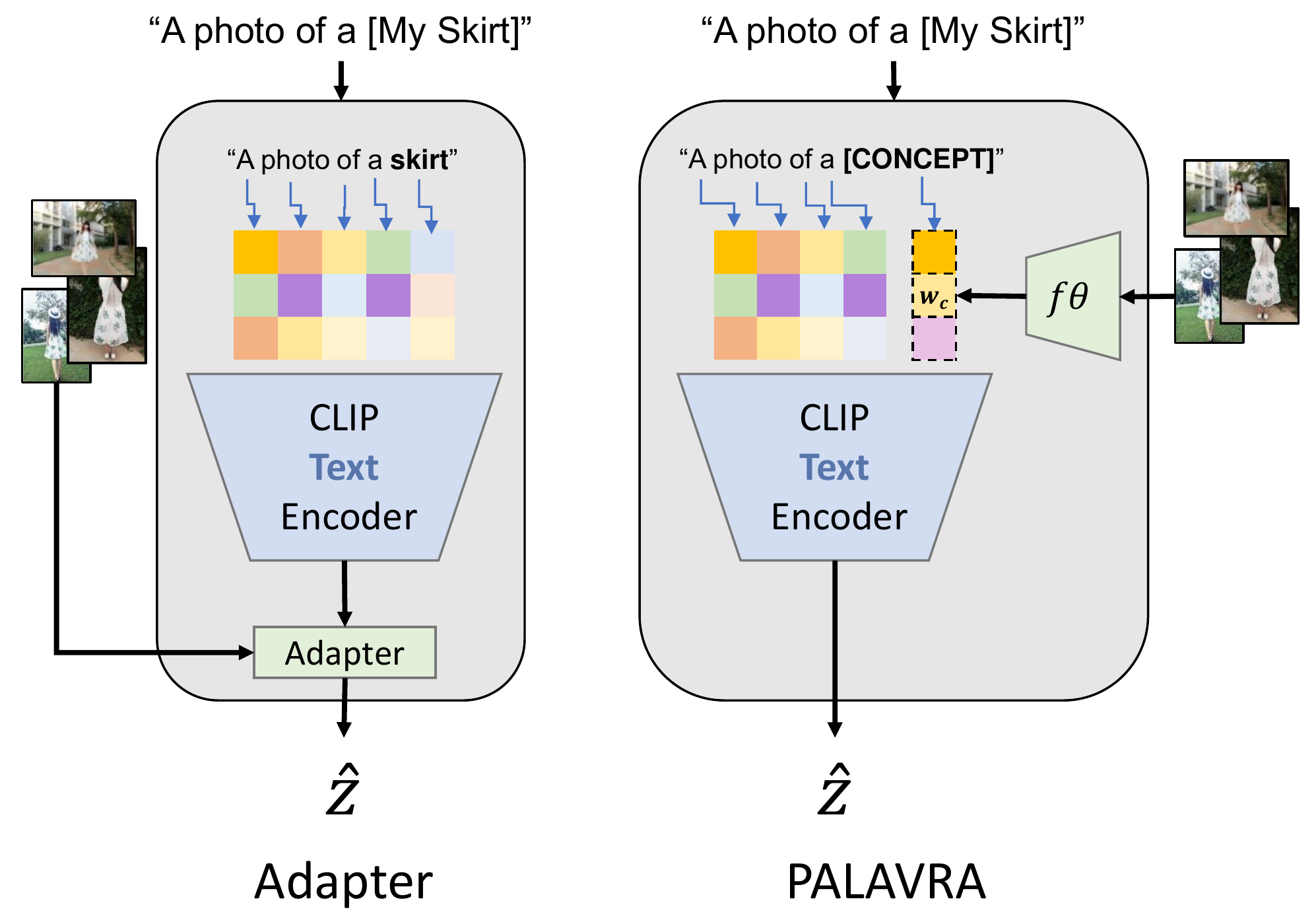} 
    \caption{Visualization of an adapter-based approach (left) and PALAVRA (right). Adapters change CLIP's output space by appending additional layers following the encoder. Our method defines new tokens in CLIP's existing input space, leaving the output space unchanged.}
    \label{fig_adapter_vs_ours}
\end{SCfigure}

Conversely, our approach does not overrides the encoder outputs. Our working hypothesis is that the text input space of a web-scale \VL model is rich enough for reasoning about new personalized concepts. We just need to find the right word embedding representation for any new personalized concept. We illustrate this architectural distinction in \figref{fig_adapter_vs_ours}.

\edit{Finally, one could fully retrain a CLIP model with the expanded vocabulary set. However, retraining CLIP requires \tildeapprox $400M$ images. Our approach is trained with a tiny fraction of that, $< 1 M$ samples, and once it is trained, different users can use it, each with their own vocabulary.}

\noindent\textbf{Notation:}
For brevity, we describe adding a single concept $c$. Adding multiple concepts can be  done iteratively. We use the notation \concept{}  to refer to a learned concept ($c$) within a textual query. $\I$ denotes the CLIP embedded image space, $\T$ the CLIP embedded textual space, $z_k = \CLIPI(\img_k)$ is the embedding of an image $\img_k$ into $\I$, and similarly $\CLIPT(S)$ is the embedding of a sentence $S$ into $\T$. Finally, $\W$ denotes the space used to embed input word tokens into CLIP.

\begin{figure}[h]
    \centering
    \includegraphics[height=185pt, trim={0cm 0cm 0cm 0cm},clip]{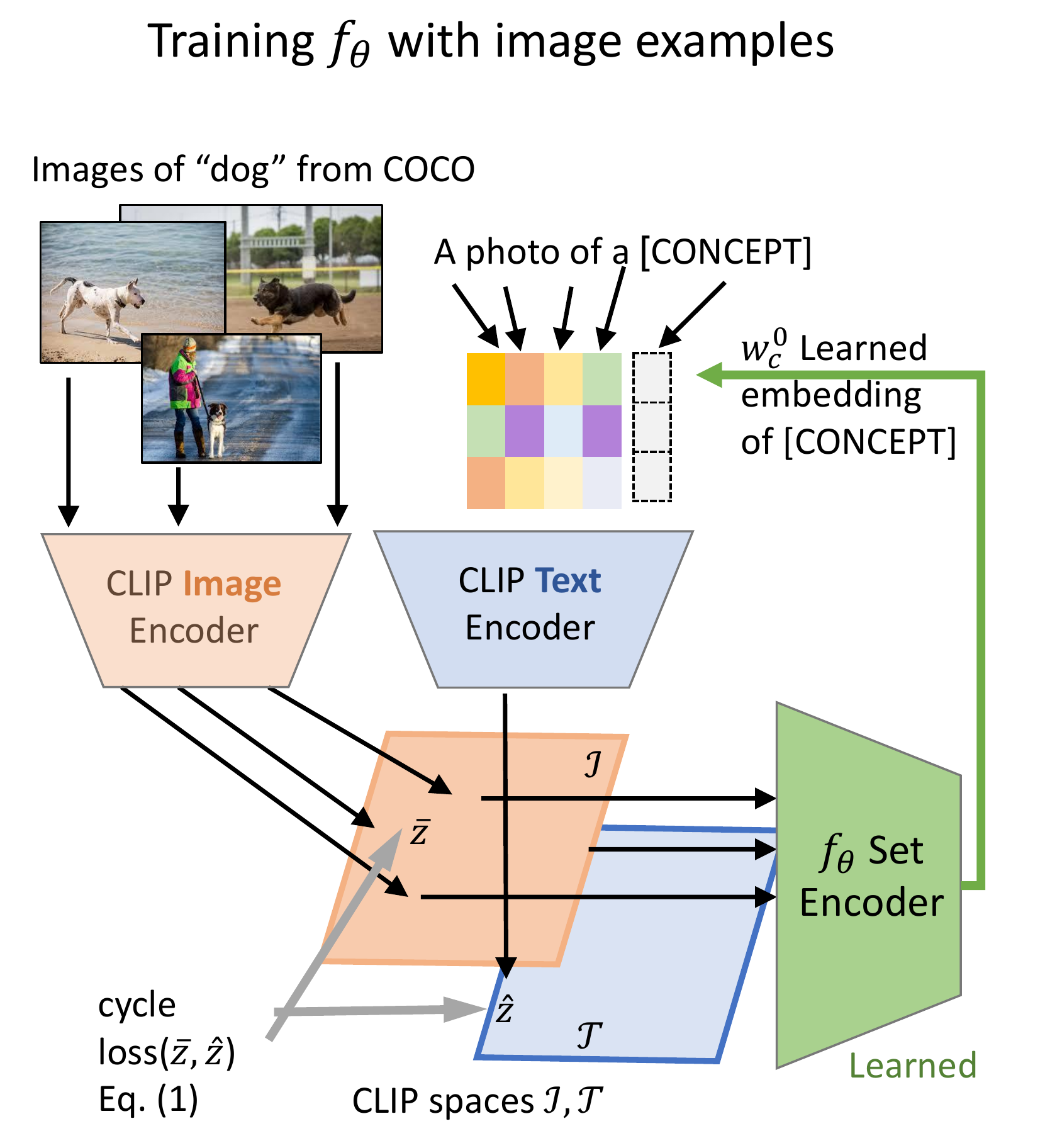} %
    \includegraphics[height=185pt, trim={0cm 0cm 0cm 0cm},clip]{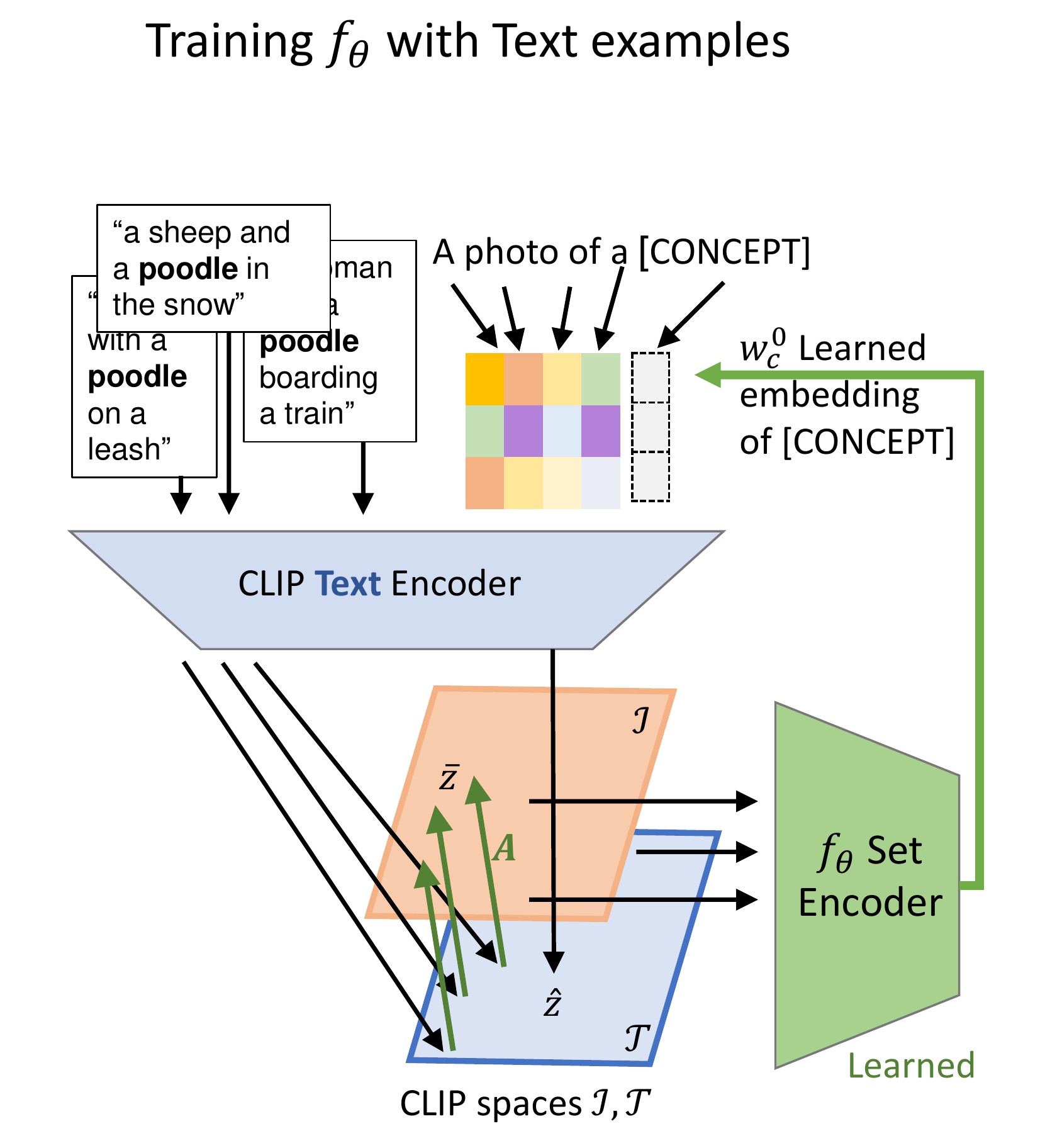} %
    \caption{\textbf{Architecture outline: Learning $\ftheta$.} We start with a \textit{large-scale-data} training step. A set encoder $\ftheta$ is trained to map CLIP-space output embeddings to a code in CLIP's input space. It is alternatingly trained with a batch of either image examples (left), or sentence examples (right) with augmented concept types. We use a cycle loss by mapping the code back to CLIP's output embedding, using a template sentence.}%
    \label{fig_ftheta}
\end{figure}

\begin{SCfigure}
    \centering
    \includegraphics[width=0.64\textwidth, trim={0.5cm 0cm 0.5cm 0cm},clip]{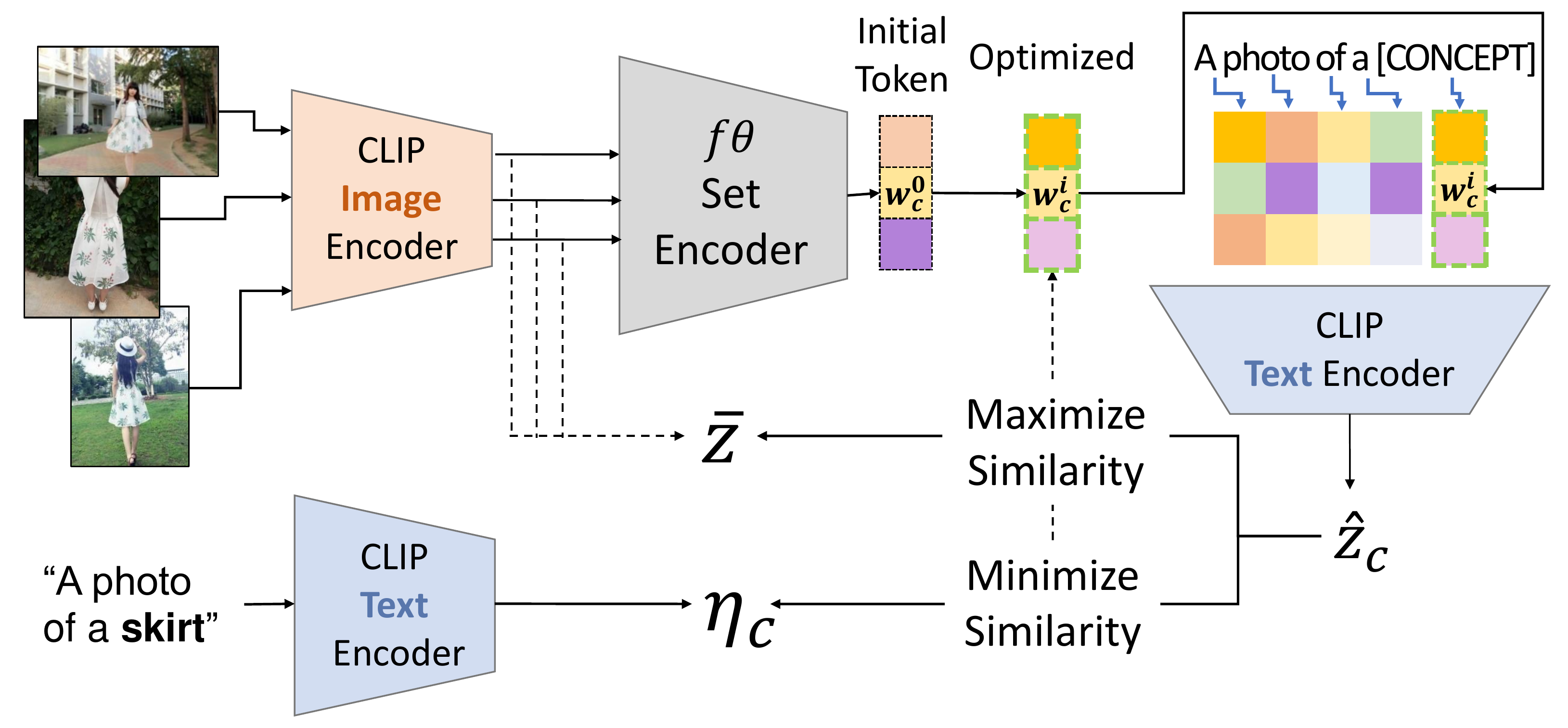} %
    \caption{\textbf{Architecture outline: Personalization.}  Given a set of examples of a personalized concept and its type (skirt), we embed them with CLIP and predict an initial code ($\w_0$) for the concept using a frozen $\ftheta$. We then further tune the code with a contrastive loss.}
    \label{fig_personalize}
\end{SCfigure}

\noindent\textbf{Architecture and Workflow:}
At a high level, our workflow has three steps.  

\textbf{(1)  Learn an inversion mapping  $\ftheta$} from a \textit{set} of points in CLIP image space $\I$ to a point in its word embedding \textit{input} space $\W$ (\figref{fig_ftheta}).
Formally, $\ftheta: \{z_k \in \I\}^{K}_{k=1} \rightarrow \W$. It is trained with \textit{non-personalized, large-scale} data.

\textbf{(2) Initial personalization} %
(\figref{fig_personalize}). Learn a word embedding $\w_c$ of a new personalized concept $c$. Thus, given a set of image examples $\img_1,...,\img_K$ we map them to CLIP image space, then map them using $\ftheta$ to obtain an initial word embedding $\w_c^0=\ftheta(\{\CLIPI(\img_k)\})$.
Formally, $\{\img_k\}_{k=1}^K \rightarrow \{\CLIPI(\img_k)\}_{k=1}^K \rightarrow \w^0_c \in\W$.

\textbf{(3) Fine-tuning.} The initial embedding $\w_c^0$ is then updated using gradient steps \edit{to maximize the similarity of the template text embeddings to the image examples, while contrasting it with an embedding of a ``super-concept''.}

Next, we describe the learning of each component in more detail.

\subsection{Learning the inversion mapping $\ftheta$}

\label{learning_ftheta}
We now describe how we learn an ``inversion" map $\ftheta$ from a set of points in CLIP space $z_1,...,z_k \in \I$, to a word embedding $\w^0_c\in\W$, where $W$ %
is the input space of the language encoder.  We base $\ftheta$'s architecture on ``Deep Sets'' \cite{zaheer2017deep}.  We now discuss the loss and how to train $\ftheta$ with two types of \textit{large-scale, non-personalized} data: images and text. See \figref{fig_ftheta}.

\textbf{A contrastive cycle loss.}
\edit{$\ftheta$ maps from CLIP space to $\w^0_c\in\W$. Then, by pairing $\w^0_c$ to the word embedding for \concept{}
we can feed $\w^0_c$ into $\CLIPT$ with a template sentence $T_c$ like ``A photo of a \concept{}''. We can then define a cycle consistency loss to match the input of $\ftheta$ with the output of $\CLIPT$ (see \figref{fig_ftheta} left)}. Specifically, let $\bar{z}_c$ be the average over samples in $\I$ from the concept $c$, $\bar{z}_c = \sum_{k=1}^K{z_k}/K$ and let
$\hat{z}_c$ be the CLIP embedding of a template sentence, $\hat{z}_c = \CLIPT(T_c)$. We wish to tune $\ftheta$ so that $\hat{z}_c$ is close to $\bar{z}_c$ for the concept $c$ and far from other concepts. We therefore define a symmetric contrastive loss for a concept $c$, with a formulation similar to SimCLR \cite{chen2020simple}: %

\begin{equation} 
\begin{split}
\label{eq_total_loss}
    \ell_{Cycle}\big({c}; \{\bar{z}_{c'}, \hat{z}_{c'} \}_{c'=1}^{C} \big)  = -\log\frac{e^{\cossim{\bar{z}_{c}}{\hat{z}_{c}}}}{\sum_{c'=1}^{C} e^{\cossim{\bar{z}_{c}}{\hat{z}_{c'}}} + \sum_{c' \neq c} e^{\cossim{\hat{z}_{c}}{\hat{z}_{c'}}}} \\
      -\text{log}\frac{e^{\cossim{\bar{z}_{c}}{\hat{z}_{c}}}}{\sum_{c'=1}^{C} e^{\cossim{\hat{z}_{c}}{\bar{z}_{c'}}} + \sum_{c' \neq c} e^{\cossim{\bar{z}_{c}}{\bar{z}_{c'}}}},
\end{split}
\end{equation}

\normalsize
\noindent where $\cossim{\hat{z}}{\bar{z}}$ denotes cosine similarity, $C$ is the number of concepts in a batch. 

\edit{We also use a regularization term $\ell_{GT}$ that maximizes the similarity of the predicted $\w^0_c$ with its ground truth. See details in the Appendix.}
\edit{Finally, the cycle loss and ground-truth regularization terms are combined with a  hyperparameter $\lambda_{gt}\geq 0$, and the total loss is $\ell_{total} = \ell_{Cycle} + \lambda_{gt} \cdot \ell_{GT}$.}

\textbf{Training $\ftheta$ with images.} We use a variant of COCO  \cite{lin2014microsoft} that extracted the subject and object from each caption as in \cite{atzmon2016learning}, and take the $1000$ most frequent concepts. In every training batch, we draw at random $C$ concepts, then draw $K$ images for each concept. We then map them to CLIP image space, yielding $\{z_k\}_1^K =\{\CLIPI(\img_k)\}_1^K$ in CLIP image space for each concept.

\textbf{Training $\ftheta$ with text.}
When training with COCO data, $\ftheta$ learns concepts that are frequent in COCO captions. However, our goal is to have $\ftheta$ generalize to widely diverse concepts.
Yet, naively training with the COCO images does not generalize well to out-of-COCO-vocabulary concepts (see \ref{sec_ablation_study}).

To generalize to out-of-vocabulary concepts, we propose synthesizing \emph{textual} descriptions with an expanded vocabulary
and embed them into the shared embedding space. Specifically, we use COCO \textit{captions} of a concept to generate additional training examples for new concepts by replacing the concept type with the most similar concept type from a large predefined vocabulary of 20K types \cite{kuznetsova2020open}, where (cosine) similarity is measured in CLIP text space. Finally, we embed the augmented captions by taking their CLIP-text feature representation (\figref{fig_ftheta}, right). %
Overall, we found that training with augmented text representation significantly improved the performance of the model (see Table \ref{tab:deep_fashion_abl}).

As in \cite{MindtheGap}, we observed that the encoding distribution of text and images does not overlap in CLIP space. As a result, training $\ftheta$ with CLIP embeddings of captions does not generalize well to image inputs. We address this problem by learning an alignment matrix $\A$ that maps CLIP representations of texts to their presumed image counterpart (\figref{fig_ftheta}, right). $\A$ is learned jointly with $\ftheta$\edit{, and is only used when learning the personalization tokens. It is not used at inference time.} Formally, a set of captions is first encoded by $\CLIPT$, then mapped to the image area of the CLIP space using $\A$ and then fed to $\ftheta$.

\subsection{\textit{Personalization}: Learn an embedding of personalized concepts}
To learn the word embedding of a personalized concept, we follow a similar process to training $\ftheta$, but instead of tuning the parameters of $\ftheta$, we optimize the actual embedding vector $\w_c$.

Specifically, let $\{\img_k\}_{k=1}^{N_c}$ be a set of input images for the new concept $c$, we (1) map them to $\I$ using CLIP encoder $\CLIPI$, (2) map to $\w^0_c$ using $\ftheta$, (3) plug the embedding $\w_c^0$ in a template sentence and (4) map to CLIP text space using $\CLIPT$. Once again, we define a contrastive cycle-consistent loss, matching the estimated text embedding of the template sentence $\hat{z}_c$, and average image embedding $\bar{z}_c$.
However, here we contrast it with the embedding of the concept \textit{type}, say ``a short sleeve top'', denoted by $\eta_c = \CLIPT(S_c)$. Since no negative image examples are provided in the personalization stage, the concept type can be viewed as a "super-concept" in the hierarchy. It allows the learning process to focus on the specific features that make the object unique from the general population of similar concept types. Similar to the SimCLR \cite{chen2020simple} loss, our loss is:

\begin{equation} \label{eq_coarse_loss}
    \ell\big(\hat{z}_c,\bar{z}_c,\eta_c \big) = -\text{log}\frac{\exp{\big( \cossim{\bar{z}_c}{\hat{z}_c} \big)}}{ %
     \exp{\big(\cossim{\bar{z}_c}{\hat{z}_c} \big)} + 2\cdot\exp{\big(\cossim{\eta_c}{\hat{z}_c} \big)}}
\end{equation}
\edit{The factor $2$ results from contrasting  $\eta_c$ with both  visual and a text embedding.}

\subsection{Inference}
Our approach expands the vocabulary of word-embedding tokens with personalized tokens, without modifying the underlying \VL model $h$. Therefore, for a given downstream task $T$ and a sentence $S$, we use the pretrained \VL model $h$ as it would have been used with $T$. But when we encounter an input sentence $S$ that includes a \concept{} token,
we apply its learned embedding $\w_c$. Also, we found that having \concept{} followed by the concept type $S_c$ is beneficial.

\section{Evaluation datasets for \PerVL } 

We created two new personalization benchmark datasets for the evaluation of PerVL. (1) We collected captions for images from DeepFashion2 \cite{DeepFashion2}, which serve as search queries in an image retrieval task. (2) We collected captions for frames from Youtube-VOS \cite{xu2018youtube}, and also collected their corresponding segmentation maps for a referring-expression segmentation task.
\begin{figure}[t]
    \centering   
    \includegraphics[width=1.0\textwidth, trim={0cm 10cm 0.1cm 1.5cm},clip]{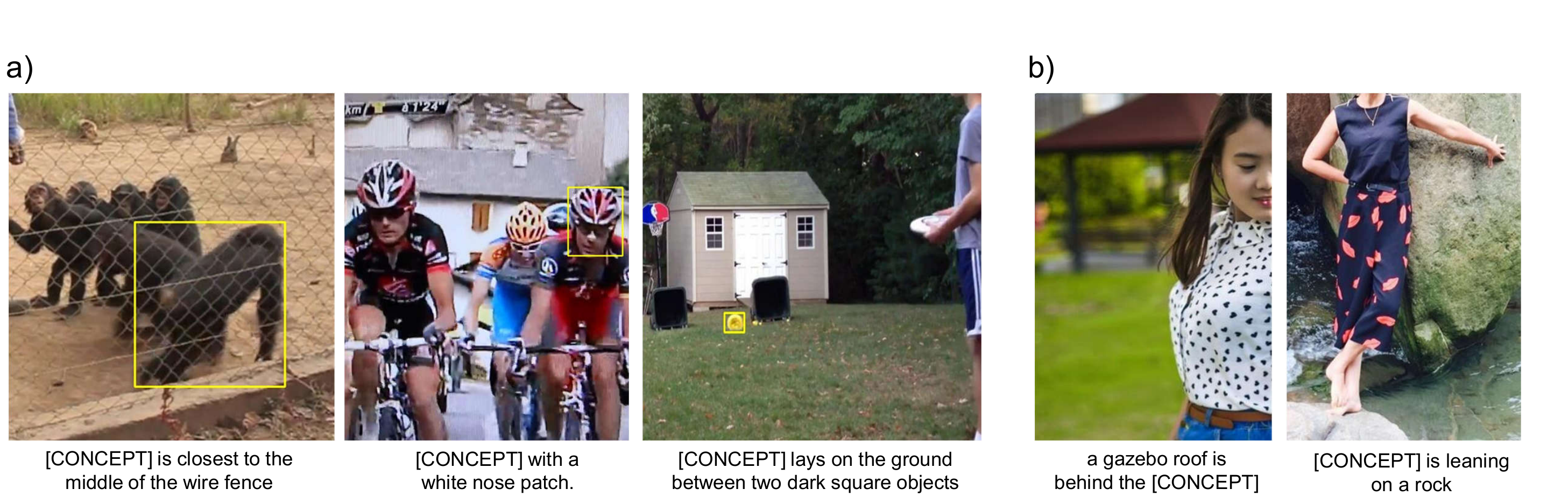}
    \caption{Examples of textual queries and evaluation images of the new benchmarks, \concept{}  %
    denotes the personalized concepts.  \textbf{(a)} YTVOS frames. The queried concept is highlighted in a yellow box. In YTVOS segmentation, one should segment the correct concept in the frame. In YTVOS retrieval, each evaluation image is a box cropped around the concept. \textbf{(b)} Deepfashion2 examples.
    }
    \label{fig_dataset_examples}
\end{figure}

\subsection{Personalized fashion-item retrieval with DeepFashion2}
We used the DeepFashion2 dataset \cite{DeepFashion2} to create an image retrieval benchmark of personalized fashion items given a textual query. It contains  photos of people wearing unique fashion items from 13 popular clothing categories, like \textit{skirt} or \textit{long-sleeve dress},  which we use as concept types. See the examples in  Fig. \ref{fig1}, \ref{fig_dataset_examples}.

We created a dataset of 1700 images from 100 unique fashion items (concepts). Each item was assigned a unique \concept{} tag.  We assigned 450 images (out of the above 1700) to an \textit{evaluation set}, and used raters to collect a textual description referring to each item appearing in the images. For instance, \textit{The \concept{}
is facing a glass store display."}. %
In Appendix  \ref{sec_deepfashion_suppl} we describe the steps we took to select context-rich items. 

\textbf{Short versus detailed captions:}
We collected two types of captions for each image. First, \textit{detailed} captions like ``\textit{White cabinets, some with open drawers, are alongside and behind the \concept{}.}''. These describe extensive context about the image and can facilitate retrieval. Second are \textit{short} captions like ``\textit{White cabinets are behind the \concept{}.}''. These pose a greater challenge, because they describe less detail, and therefore are more ambiguous.

Finally, we randomly split the data to 50 val. concepts and 50 test concepts.

\subsection{Youtube-VOS for personalized retrieval and segmentation.}
We created an image segmentation benchmark of personalized visual concepts given a textual query using Youtube-VOS (YTVOS)~\cite{xu2018youtube}. YTVOS is a dataset for instance segmentation in video, which includes 4000+ videos, 90+ categories, and 7800+ unique object instances. To transform it to an image personalization benchmark, we take the last frame of each video (scene) for evaluation and the object instances that appear in the frame as the target concepts. Earlier frames are used as candidate frames for training. See the examples in \figref{fig_dataset_examples} (left).

This benchmark is challenging as it contains ambiguities about both the textual queries and the appearance of the personalized concept. Hence, only a model that is successful in both personalization and image-text reasoning can succeed in this task. For that, we only select videos such that their object concept appears at least twice in an evaluation frame. 

Finally, we annotated the instances in the evaluation frame with captions using AMT. We instructed the AMT workers to concisely describe what makes a specific entity distinct, compared to similar entities in the image. We provide more details and examples in Appendix \ref{app:ytvos_details}.

In total, this benchmark includes $\tildeapprox\!$500 unique personalized concepts, with $\tildeapprox\!$6300 training samples. For evaluation, we split according to unique scenes (videos), resulting in 246 validation concepts and 251 test concepts.

\textbf{Personalized image retrieval:}
We also created an image retrieval variant of YTVOS. We extracted a set of images that correspond to the AMT captions collected for segmentation. Every image in the retrieval set was extracted from a wide box cropped around every instance in each evaluation frame. The goal is to retrieve the image of the correct instance given its textual query. Compared to the segmentation task, there are fewer distractors from the same scene for every instance, since not all instances were labeled in the data, but there are many more distractors coming from different scenes.

\section{Experiments}
\label{sec_experiments}
We tested PALAVRA with two \PerVL benchmarks and compared with several leading baselines (Sec.\ref{sec_retrieval_results},\ref{sec_segmentation_results}). We then study in greater depth the properties of PALAVRA, by an ablation study (see \ref{sec_ablation_study}). All design decisions and hyperparameter tuning were performed on the validation set to avoid overfitting the test set. The experiments were carried out on NVIDIA V100 and A100 GPU. We provide additional implementation details and results in Appendix \ref{app:implementation_detais},\ref{sec_additional_results}.

\begin{figure*}[t]
\begin{tabular}{ll}
\includegraphics[height=0.28\textwidth, trim={0cm 0cm 19.5cm 0cm},clip]{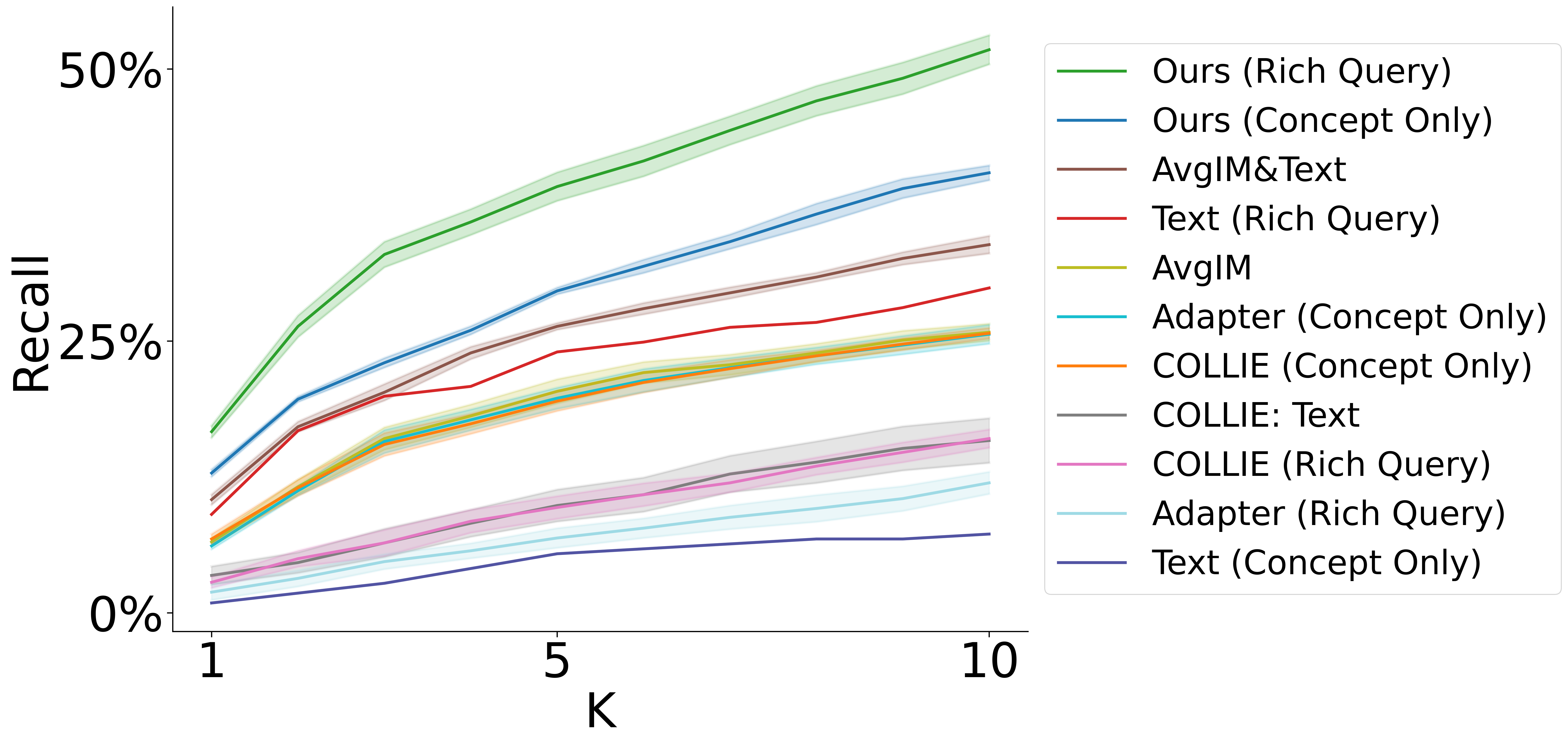} %
\includegraphics[height=0.28\textwidth, trim={39cm 6cm 0.6cm 2cm},clip]{figures/deepfashion_4pages.png} %

\includegraphics[height=0.28\textwidth, trim={0cm 0cm 0.6cm 0cm},clip]{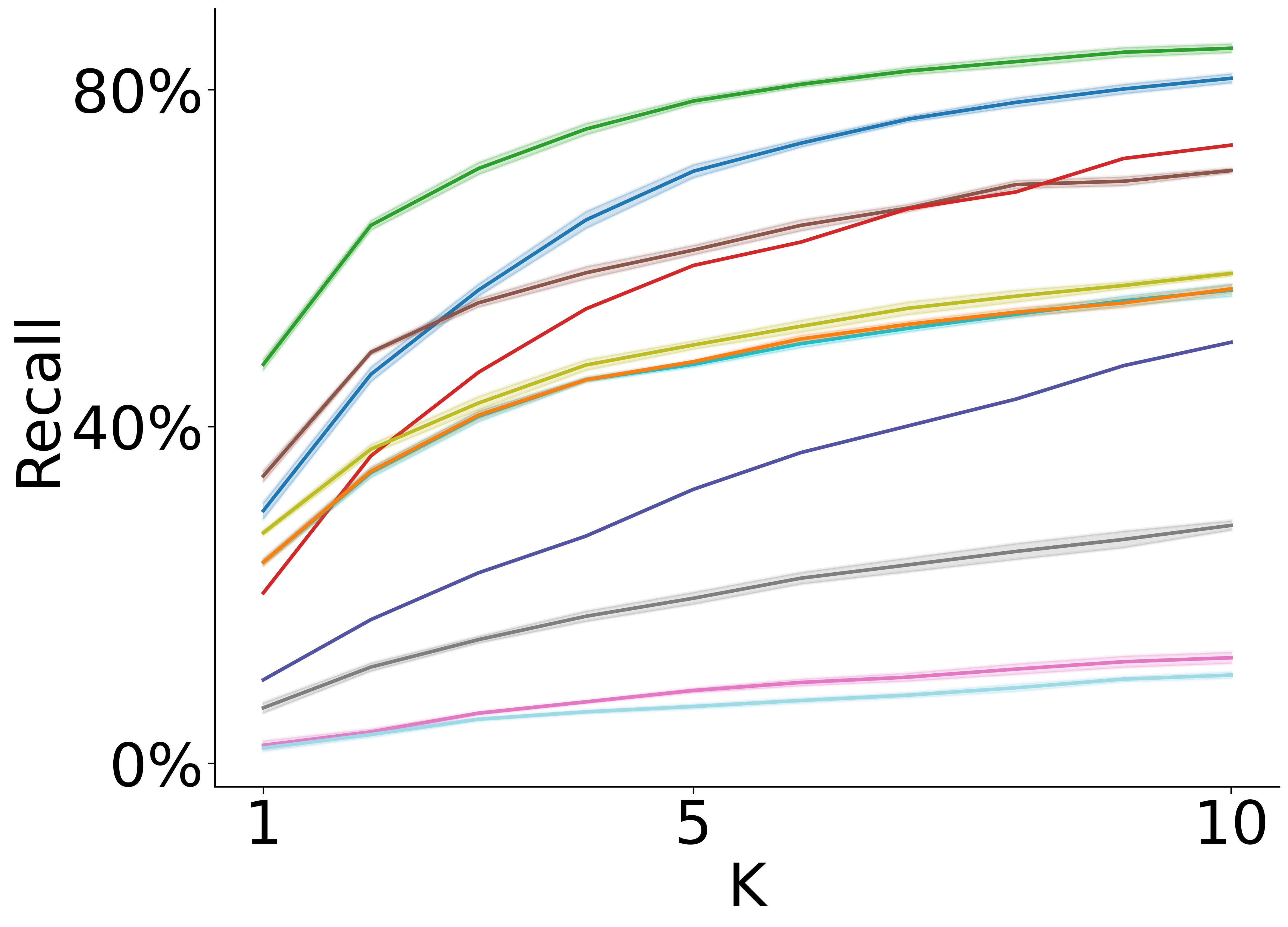}&
\end{tabular}
\caption{Recall at K for our approach and baselines.  DeepFashion2 (left), Youtube-VOS (right),
    PALAVRA achieves the highest rates for all retrieval metrics. On both benchmarks and metrics it achieves a significant improvement  compared with ``AvgIm\&Text'', which is the strongest baseline.
    The experiments with ``Concept-only query''  demonstrate that the information in the textual query is essential for telling the target image from distractors, since the performance of both PALAVRA and ``Text Only'' substantially degrades with ``Concept-only'' queries. }
\label{fig_recall_at_k}
\end{figure*}

\subsection{Personalized image retrieval with a textual query}
\label{sec_retrieval_results}
The objective of this task is to retrieve the correct image given a text query that includes the new concept (\figref{fig1}, top-right). We use AMT captions as textual queries describing a single image from the dataset. %
The challenge in this setup is to overcome two types of distractors. (a) visually similar distractors: images of the same personalized concept but in a different context than the context described by the textual query (e.g. the two instances of ``my favorite skirt" in \figref{fig1}), (b) semantically similar distractors: images which include an item of a similar concept type (e.g. ``a skirt''), in a similar context as described in the textual query.

To rank images according to a textual query, we rank images according to their cosine similarity with the embedded text query.

\noindent\textbf{Compared methods. }\label{sec_comp_methods_retrieval} We compare our approach \textbf{PALAVRA}, with 5 baselines and their variants: \textbf{Text Only}: score an image-query pair using CLIP embedding of the text query $\CLIPT(S)$. Using the concept type for \concept{} instead of its learned word embedding.
\textbf{AvgIM}: \edit{Ignoring the text query and replace it} by the average over the embedding of its concept training images. %
\edit{This is equivalent to the FSL baseline in \cite{chen2021meta}.%
}
\textbf{IM\&Text}: Represent the query as the average between \hfill \textit{AvgIM} \hfill and \textit{Text}. \hfill  \textbf{Random}: Test \hfill images \hfill are \hfill ranked \hfill in  \hfill  random  \hfill order. 

\textbf{COLLIE} \cite{skantze2021collie}:
 Learn an adapter module over the output of CLIP text encoder, with an additional scaler function $Scaler\big(\CLIPT(S)\big) \in [0,1]$ that softly applies the adapter layer. COLLIE is closest to our method, because it may preserve some capabilities of the underlying pretrained model, when $Scaler(\cdot)=0$.
\textbf{Adapter}: As in COLLIE, but replace the scaler with a constant value of $1$, 
 making the ``Adapter'' layer always active.
\textbf{COLLIE:Text}: COLLIE, when the text query uses the concept type for \concept{}, rather than the trained concept.

\begin{table}[t]
\centering
    \begin{tabular}{l|c|c}

    \toprule
        {} &    \multicolumn{1}{l|}{DeepFashion2}       &    \multicolumn{1}{l}{YTVOS} \\
    {} &         MRR &         MRR \\
    \midrule
    Random         &   2.9 $\pm$ 0.4 &   2.8 $\pm$ 0.2 \\
    \toprule
    {\textbf{Concept-only query}} &&\\
    Text Only    &  4.2 $\pm$ 0.0  &  21.5 $\pm$ 0.0 \\
    Adapter        &  13.4 $\pm$ 0.5 &  35.5 $\pm$ 0.3 \\
    COLLIE         &  13.8 $\pm$ 0.5 &  35.6 $\pm$ 0.3 \\
    AvgIm          &  13.8 $\pm$ 0.5 &  38.2 $\pm$ 0.3 \\
    PALAVRA (Ours) &  19.4 $\pm$ 0.6 &  53.4 $\pm$ 0.8 \\    %
    \toprule
    {\textbf{Rich query}}&&\\
    Adapter        &   5.9 $\pm$ 0.7 &   5.3 $\pm$ 0.3 \\
    COLLIE         &   7.9 $\pm$ 0.7 &   6.2 $\pm$ 0.3 \\
    COLLIE: Text    &  8.0 $\pm$ 1.0 &   7.2 $\pm$ 0.3 \\
    Text Only    &  17.6 $\pm$ 0.0 &  37.6 $\pm$ 0.0 \\
    AvgIm+Text     &  18.8 $\pm$ 0.4 &  47.2 $\pm$ 0.3 \\
    \midrule
    PALAVRA w.o. tuning   &  22.1 $\pm$ 0.2 &  47.1 $\pm$ 0.8 \\
    PALAVRA (Ours) &  \textbf{28.4 $\pm$ 0.7} &  \textbf{61.2 $\pm$ 0.4} \\
    \bottomrule
    
    \end{tabular}

    \caption{MRR retrieval metrics.}
    \label{table_main_retrieval}
\end{table}

\noindent\textbf{Evaluation metrics and queries.} %
For image retrieval, we report two metrics (1) \textbf{Recall at K}: The rate of successful retrievals among the top-K scoring images. (2) \textbf{MRR} (Mean Reciprocal Rank): Average of 1 divided by the rank of the correct image. Errors denote the standard error of the mean (SEM) across 5 model initialization seeds.
We use two types of queries: \textbf{(1) Rich query} uses the free-formed text annotated by AMT workers: ``\concept{} is leaning on a rock''. \textbf{(2) Concept-only queries} overrides the ``Rich query`` by a template that focuses only on the concept: ``A photo of a \concept{}''. Note that the baseline ``AvgIM''  is more related to the ``Concept-only query'' because the rich query embedding is overridden by the average embedding of the training examples.

\noindent\textbf{Retrieval Results.}
Table \ref{table_main_retrieval} and \figref{fig_recall_at_k} describe the retrieval rates of PALAVRA and the compared methods when using challenging \textit{short} captions as our \textbf{Rich Queries}. We report the retrieval rates with \textit{detailed} captions in the appendix. We note that both \textit{short} queries and \textit{detailed} queries are rich queries, containing known concepts in addition
to the personalized ones. The \textit{detailed} version possibly contains more of them. PALAVRA achieves  the highest rate in all the  retrieval metrics. On both benchmarks and metrics, it achieves significant improvement (between $\tildeapprox30\%$ to $\tildeapprox50\%$) compared with ``AvgIm+Text'', which is the strongest baseline.

Comparing the results of ``Concept-only query'' with the ``Rich query'' results  demonstrate that: (1) Information in the ``Rich query'' is essential for retrieval.   (2) Adapter baselines (Adapter \& COLLIE) improve over vanilla CLIP when only the concept is queried.

 Their performance degrades when using the ``Rich query''. This happens as the adaptation layer trained for the personalized \concept{} does not perform well with free-form text it has not seen during training. %
In fact, we find that Adapter and COLLIE are even sensitive to the prompt \textit{prefix} of the query. When changing their prefix to a prefix not used in training, their performance substantially degrades. We report this finding quantitatively in Appendix \ref{sec_additional_results}.

\begin{figure}[htbp]
    \centering
    \begin{tabular}{ccc}
        \begin{overpic}[width=0.405\linewidth, trim={0cm 0cm 1.2cm 1.4cm},clip]{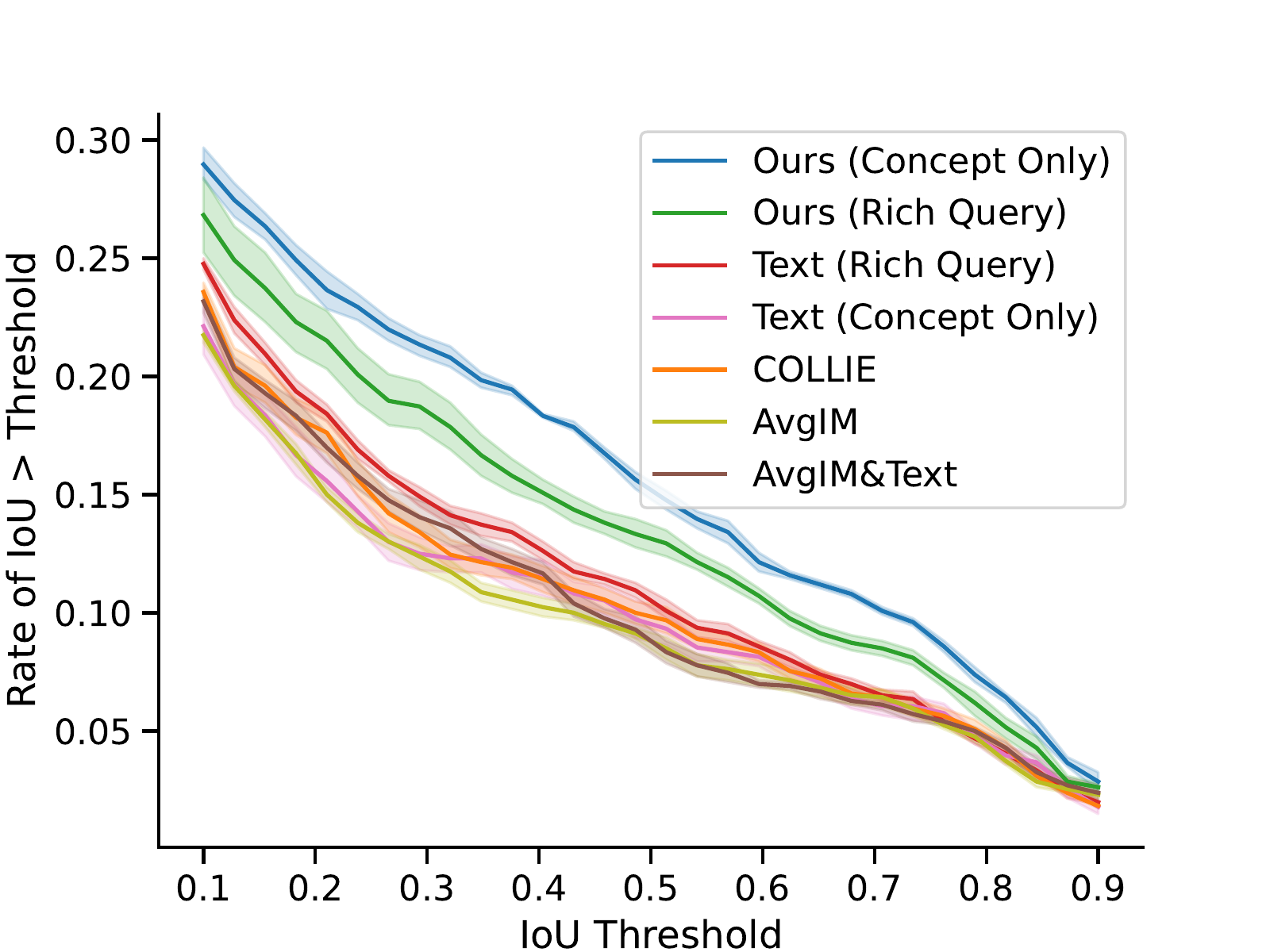} %
        \put (4,4) {(a)}
        \end{overpic} &
        \begin{overpic}[width=0.285\linewidth, trim={0cm 0cm 0.7cm 1.2cm},clip]{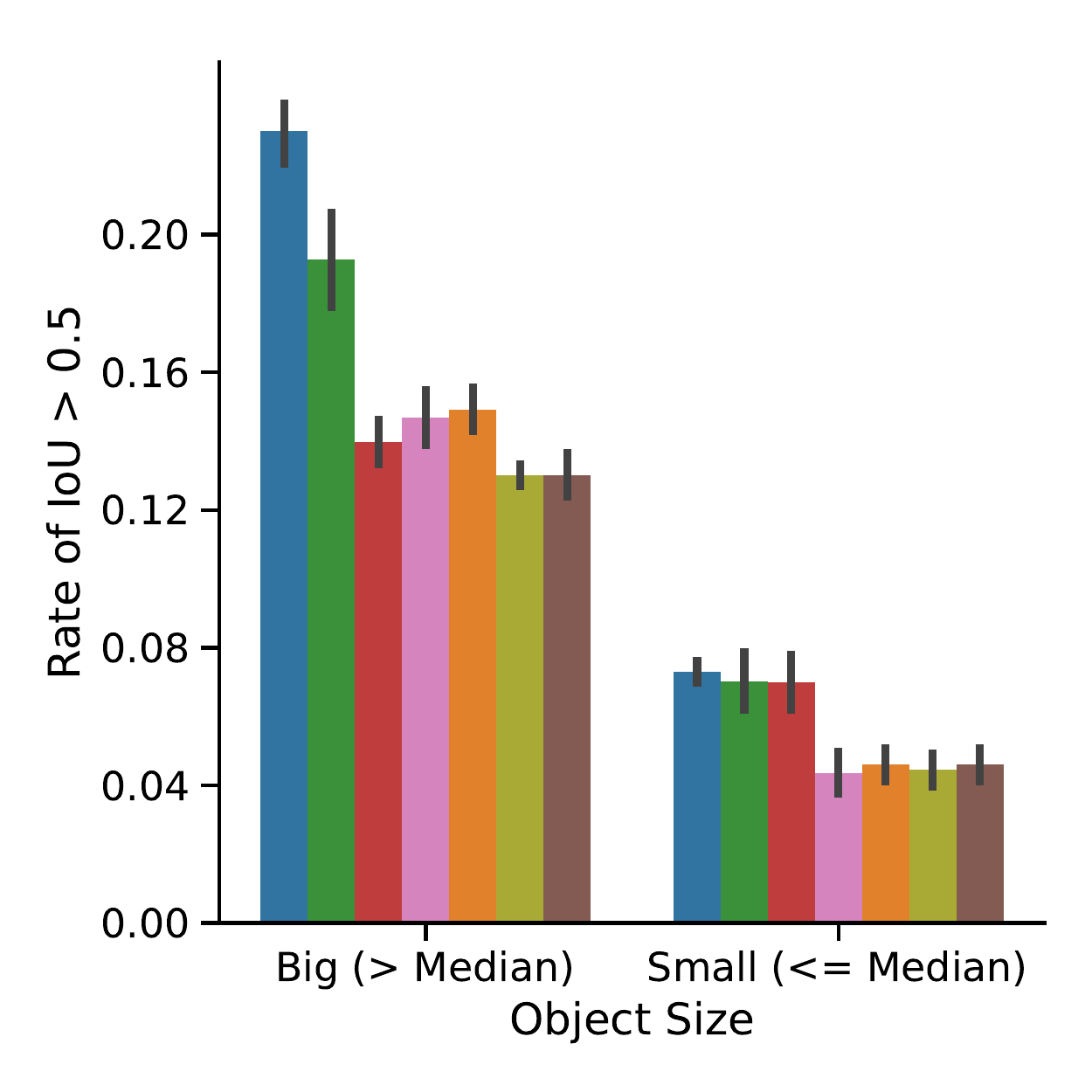}  %
        \put (4,5) {(b)}
        \end{overpic} &
        \begin{overpic}[width=0.285\linewidth, trim={0cm 0cm 0.7cm 1.2cm},clip]{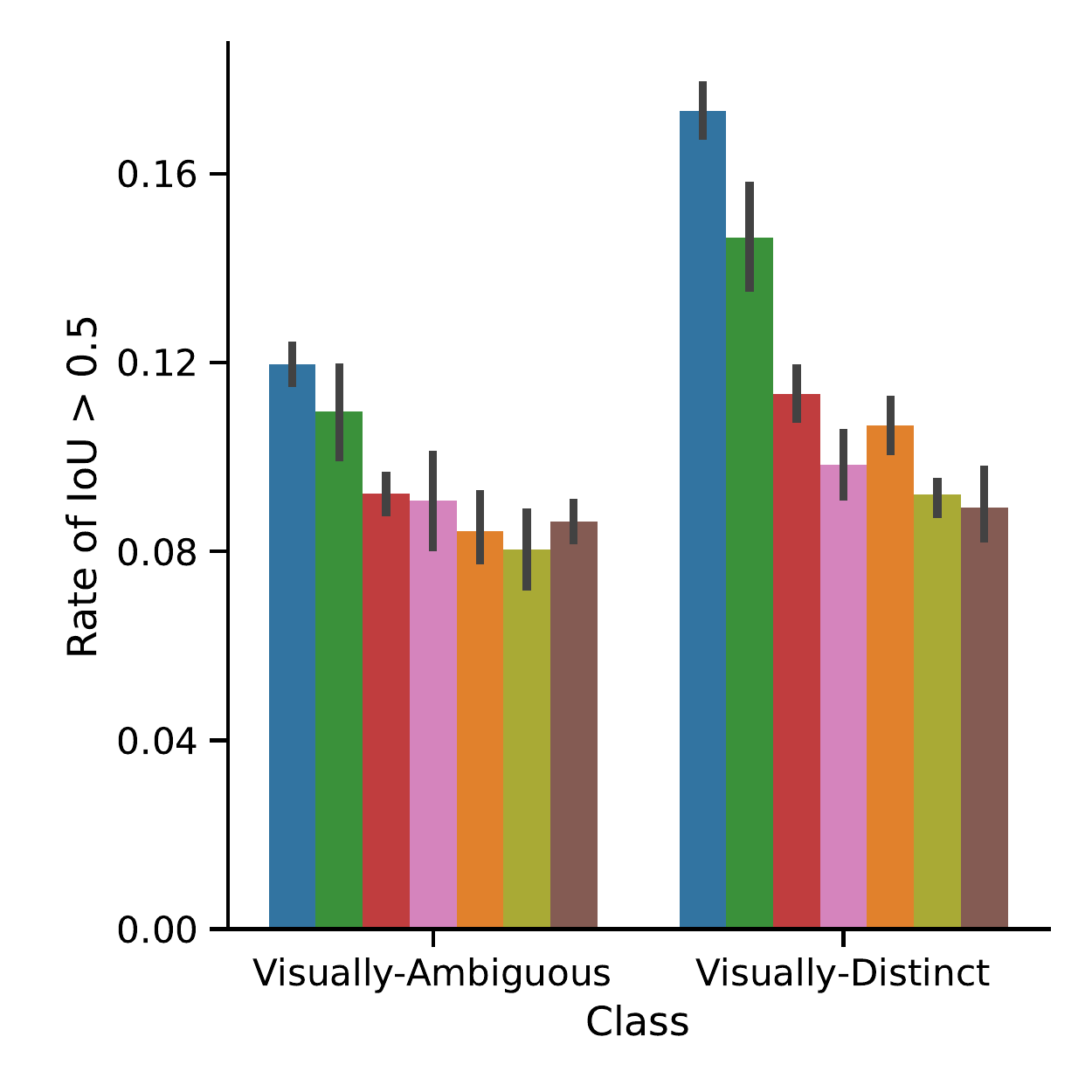}  %
        \put (4,5) {(c)}
        \end{overpic} %

    \end{tabular}
    \caption{Segmentation results on YTVOS. \textbf{(a)} Percent of images where IoU values exceeded a threshold, as a function of the threshold. Our approach dominates across the full range \textbf{(b, c)} Investigating model robustness under $2$ levels of task complexity. We consider two scenarios that influence difficulty: Object size (b), and intra-class class ambiguity (c). When a clear visual signal is available (large objects, high intra-class visual variance), our model significantly outperforms the alternatives. Even in more challenging scenarios, our model still leverages textual descriptions, mitigating the loss of quality seen in models that ignore or corrupt CLIP's embedding space. }
    \label{fig:seg_results}
\end{figure}

\subsection{Semantic segmentation with a textual query}
\label{sec_segmentation_results}
The second downstream task, aims to segment an instance of a personalized concept in an image, based on a textual query that refers to the concept (\figref{fig1}, right bottom).  
The challenge here is to overcome two types of distractors: First, \textit{visual distractors} that look similar to the concept and can be disambiguated with the text query. Second, \textit{semantic distractors} that include a concept of a similar type (e.g. another type of a ``toy wagon''), but %
CLIP has difficulty to resolve using just the  concept type, like ``an elephant on a toy wagon''. %

Here we investigate the performance of PALAVRA and baseline models on YTVOS dataset using a recent CLIP-based semantic segmentation  \cite{zabari2021semantic}. In brief,  \cite{zabari2021semantic} creates a set of query-driven relevance maps for the image, coupled with transformer interpretability methods~\cite{Chefer_2021_CVPR}.  The maps then serve as pseudolabels for single-image semantic segmentation~\cite{reviving2021}.
 
\noindent\textbf{Compared methods.}
\label{sec_comp_methods_sseg}
We compare PALAVRA with a set of baselines in two setups: ``Rich query'' and ``Concept-only query'', as described in  \secref{sec_retrieval_results}. 
\textbf{(1) Text (CLIP)}, using both the rich and concept only queries, \textbf{(2) AvgIM}, \textbf{(3) IM\&Text} and \textbf{ (4) COLLIE}. All baselines are described in \secref{sec_comp_methods_retrieval}. 

\noindent\textbf{Evaluation metrics.}
We calculate the intersection-over-union (IOU) between the predicted segment and the ground-truth segment. We report the \textbf{Rate of IOU $>$ threshold}, which is the fraction of segments with IOU $>$ threshold. Error bars denote the standard error of the mean (SEM) across 5 model seeds.

\noindent\textbf{Semantic segmentation results}

Fig. \ref{fig:seg_results}a shows the percent of test-set images for which IoU exceeds a given threshold. Our model consistently outperforms the baselines with wide margins (e.g. a $44.69\%$ improvement over the best competitor at an IoU threshold of $0.5$).
These results demonstrate that a personalized prompt can extend even to localized image tasks. Moreover, as our method only expands CLIP's input space - it can be readily integrated with existing models for downstream tasks. 

Surprisingly, our method performs better when using the concept-only query. We hypothesize that this is a result of CLIP's difficulty in reasoning over complex referring expressions: By mentioning the context within the text, CLIP's attention is diverted to the context itself which leads to false negatives.  
\edit{In contrast, in retrieval, context  objects rarely appear in ``negative" (non-target) images. Since they appear in the target image, they actually help to detect the correct image. Appendix \ref{app:rich_mask_vis} provides quantitative evidence supporting this hypothesis.}

We further examine cases where we expect existing methods to falter. In Fig. \ref{fig:seg_results}b, we examine a scenario where objects are small, so their crops may not provide sufficient signal to the CLIP image encoder. Indeed, when for objects below the median size, segmentation fares worse in general, and image based-methods suffer even more. Our method, however, can rely on the signal derived from the text and degrades less. 
Fig. \ref{fig:seg_results}c examines a scenario in which objects in a scene are less visually distinct. We divide the evaluation set to object which are usually visually distinct and images which often contain visually ambiguous objects, where a few instances of the same object appear in the same image. In practice, we split to animal and non-animal categories, as animals are mostly visually ambiguous (e.g. \figref{fig_dataset_examples} left).
Our model substantially outperforms the baselines when the concepts are visually distinct and also improves when the concepts are mostly ambiguous.

In the Appendix \ref{sec_additional_results} we provide and discuss qualitative segmentation results.

\section{Ablation Study}
\label{sec_ablation_study}
To understand how various components of our approach contribute to its performance, we conducted an ablation study. We report validation and test metrics for DeepFashion2 and YTVOS.

We first ablate model components that affect training $\ftheta$. We report the results without fine-tuning, to reveal how they affect the training of the set encoder. We call this stage \textbf{``no tuning''}.

Then we compare components that affect the fine-tuning stage. We call this stage \textbf{``with tuning''}. Specifically, we compare the following components:
\begin{enumerate}
    \item \textbf{PALAVRA} is our approach described in \secref{sec:methods}. We tested it both \textbf{with tuning} and \textbf{no tuning}.
    \item \textbf{no text augment} shows the results of  $\ftheta$ trained only with visual concepts that exist in the MS-COCO vocabulary, and without using the extended vocabulary for augmentation.
    \item \textbf{Only $\mathbf{\ell_{GT}}$} does not use the cycle loss for training $\ftheta$ (see Eq. \ref{eq_total_loss}).
    \item \textbf{Only $\mathbf{\ell_{Cycle}}$} does not use the ground-truth regularization term for training $\ftheta$ (see Eq. \ref{eq_total_loss}).
    \item \textbf{Only tuning} initializes $\w^0_c$ randomly, instead of using the prediction made by the set encoder $\ftheta$.
    \item \textbf{no alignment} shows the performance of our method when replacing the alignment matrix $\A$ by an identity mapping.
\end{enumerate}

Table \ref{tab:deep_fashion_abl} shows the results of the ablation experiments.
Several points worth discussing. First, PALAVRA without tuning improves by $25\%$ compared to ``Text Only'' (in Table \ref{table_main_retrieval}), for both DeepFashion2 ($22.1$ vs. $17.6$) and YTVOS ($47.1$ vs. $37.6$). This result indicates that $\ftheta$ learns to predict the word embeddings of visual concepts, and these concepts are better than using their vanilla CLIP embeddings.

\begin{table}[t]
\centering
\begin{tabular}{l|ll|ll}
    \toprule
    {DeepFashion2} &    \multicolumn{2}{l|}{Validation}       &    \multicolumn{2}{l}{Test} \\
    \toprule
    {} &       MRR &     Recall@5 &       MRR &  Recall@5 \\
    \midrule
     {\textbf{no tuning}} &&\\
PALAVRA (Ours) &  \textbf{26.9 $\pm$ 0.2} &  \textbf{35.9 $\pm$ 0.2} &  \textbf{22.1 $\pm$ 0.2} &  \textbf{29.6 $\pm$ 0.3} \\
 no text augment  &  21.8 $\pm$ 1.9 &  29.2 $\pm$ 2.3 &  19.1 $\pm$ 0.2 &  25.7 $\pm$ 0.3 \\
Only $\ell_{GT}$ &  23.3 $\pm$ 0.4 &  31.9 $\pm$ 0.5 &  19.2 $\pm$ 0.5 &  25.1 $\pm$ 0.8 \\
Only $\ell_{Cycle}$   &  19.3 $\pm$ 0.5 &  26.8 $\pm$ 0.8 &  16.1 $\pm$ 0.4 &  21.6 $\pm$ 0.8 \\
          \toprule
   {\textbf{with tuning}} &&\\

    PALAVRA (Ours)  &  \textbf{36.2 $\pm$ 1.3} & \textbf{53.7 $\pm$ 2.0}  & \textbf{28.4 $\pm$ 0.7} & \textbf{39.2 $\pm$ 1.3} \\
    Only tuning  &  32.1 $\pm$ 0.6 &  44.1 $\pm$ 0.7 &  27.5 $\pm$ 1.0 &  37.9 $\pm$ 1.8 \\
    no alignment  &  32.9 $\pm$ 0.4 &  47.8 $\pm$ 1.1 &  26.3 $\pm$ 0.2 &  36.9 $\pm$ 1.6 \\
    \bottomrule
\end{tabular}
\vspace{10pt}
\begin{tabular}{l|ll|ll}
    \toprule
        {YTVOS} &    \multicolumn{2}{l|}{Validation}       &    \multicolumn{2}{l}{Test} \\
    \midrule
    {} &       MRR &  Recall@5 &    MRR & Recall@5 \\
    \midrule
     {\textbf{no tuning}} &&\\
    PALAVRA (Ours)  &  \textbf{47.3 $\pm$ 0.5} &  \textbf{68.5 $\pm$ 0.5} &  \textbf{47.1 $\pm$ 0.8} &  \textbf{70.3 $\pm$ 0.8 }\\
    no text augment  &  45.0 $\pm$ 0.3 &  63.8 $\pm$ 0.5 &  44.4 $\pm$ 0.3 &  65.6 $\pm$ 0.4 \\
    Only $\ell_{GT}$ &  40.8 $\pm$ 0.8 &  59.3 $\pm$ 1.3 &  41.4 $\pm$ 0.2 &  62.0 $\pm$ 0.1 \\
    Only $\ell_{Cycle}$    &  35.5 $\pm$ 1.1 &  50.8 $\pm$ 2.4 &  37.3 $\pm$ 1.1 &  55.8 $\pm$ 2.1 \\
    \midrule
   {\textbf{with tuning}} &&\\
    PALAVRA (Ours)   &  \textbf{59.0 $\pm$ 0.8} &  \textbf{76.2 $\pm$ 1.1} &  \textbf{61.2 $\pm$ 0.4} &  \textbf{78.7 $\pm$ 0.4} \\
    Only tuning   &  57.3 $\pm$ 0.9 &  76.1 $\pm$ 0.9 &  57.8 $\pm$ 0.3 &  77.1 $\pm$ 0.8 \\
    no alignment &  56.5 $\pm$ 0.7 &  74.1 $\pm$ 0.3 &  58.1 $\pm$ 0.3 &  75.2 $\pm$ 0.9 \\
    \bottomrule
    \end{tabular}
    \caption{Ablation study: Highlighting the importance of various components. See the text for a full description of each setting and an analysis of the results. 
    }
    \label{tab:deep_fashion_abl}
    \vspace{-20pt}
\end{table}

Next, we find that text augmentation with extended vocabulary (\secref{learning_ftheta}) improves concept learning with $\ftheta$. It yields an improvement of $\tildeapprox 16\%$ for DeepFashion2 ($22.1$ vs. $19.1$) and $\tildeapprox 6\%$ for YTVOS ($47.1$ vs. $44.4$).

Combining a cycle loss with the ground truth (GT) regularization term is effective. When combined with the GT regularization term, the cycle loss improves by $\tildeapprox 16\%$ for DeepFashion2 ($22.1$ vs $19.2$)  $\tildeapprox 14\%$ for YTVOS ($47.1$ vs $41.4$). However, when the GT regularization term is deactivated and only the cycle loss is used, $\ftheta$ fails to generalize ($16.1$ in DeepFashion2 and $37.3$ in YTVOS).
We hypothesize that this effect is similar to the effect observed with inversion to the latent space of GANs~\cite{tov2021designing}.
There, inversions into sparse regions of the latent space can better satisfy a cyclic reconstruction loss, but they behave poorly under interpolation. Our $\ftheta$ could similarly learn to invert into sparse regions of CLIP's input space. By adding the GT regularization term, our inversions are encouraged to reside in better-behaved regions of the input space, namely those observed during CLIP's training. In these regions, the semantics of the latent space hold better and the model can better generalize. %

When $\ftheta$ is replaced by a random initialization, the performance degrades by $\tildeapprox 6\%$ for YTVOS ($57.1$ vs $61.2$) and $\tildeapprox 3\%$ for DeepFashion2 ($27.5$ vs $28.4$). Showing the synergy between the two personalization steps.

Finally, integrating the alignment matrix $\A$ showed an improvement of  $\tildeapprox 8\%$ for DeepFashion2 ($28.4$ vs. $26.3$) and $\tildeapprox 5\%$ for YTVOS ($61.2$ vs. $58.1$).

\section{Discussion}

We described an approach to leverage large pre-trained \VL models like CLIP, for learning a representation of new ``personal" classes from a handful of samples. Our key idea is to expand the input space of \VL models by finding a representation of the new concept. The extended model can then be used for \VL tasks with a rich language that ``understands" both novel and known concepts. A limitation of the approach is that it suffers from the limitations of the underlying \VL model. For instance, CLIP struggles with understanding spatial relations within a photo \cite{liu2021learning}, and extended representations based on CLIP suffer from the same problem. We expect that our approach can be extended to other \VL models. See an example in Appendix \ref{approch_extension}.

To conclude, we hope that the method presented in this paper will pave the way to using pretrained models in problems that involve user-specific concepts, like home robotics and organizing personal data.

\clearpage
\bibliographystyle{splncs04}
\bibliography{egbib}
\clearpage
\appendix
\renewcommand\thefigure{A.\arabic{figure}}    
\renewcommand\thetable{A.\arabic{table}}   
\renewcommand\thesection{\Alph{section}}   
\renewcommand{\theequation}{A.\arabic{equation}}
\setcounter{figure}{0}  
\setcounter{table}{0}  
\setcounter{section}{0}  

{\LARGE{Appendix}}

\section{Implementation details}
\label{app:implementation_detais}

\textbf{Number of training samples:} The results reported in the main text used 5 training samples for each concept in the retrieval experiments, and 10 training samples for each concept in the segmentation experiments. Below, in \secref{sec_additional_results}, we provide additional results that sweep over the number of training samples.

\vspace{15pt}\noindent\textbf{Cycle loss prompts:}
We use multiple prompts for querying the concept with the cycle loss. In each epoch, we selected a template at random from the following list of prompts.

\noindent{``This is a photo of a \concept{}", %
    ``This photo contains a \concept{}", %
    ``A photo of a \concept{}", %
    ``This is an illustrations of a \concept{}", %
    ``This illustrations contains a \concept{}", %
    ``An illustrations of a \concept{}", %
    ``This is a sketch of a \concept{}", %
    ``This sketch contains a \concept{}", %
    ``A sketch of a \concept{}", %
    ``This is a diagram of a \concept{}", %
    ``This diagram contains a \concept{}", %
    ``A diagram of a \concept{}", %
    ``A \concept{}", %
    ``We see a \concept{}", %
    ``\concept{}", %
    ``We see a \concept{} in this photo", %
    ``We see a \concept{} in this image", %
    ``We see a \concept{} in this illustration", %
    ``We see a \concept{} photo", %
    ``We see a \concept{} image", %
    ``We see a \concept{} illustration", %
    ``\concept{} photo", %
    ``\concept{} image", %
    ``\concept{} illustration".}

\vspace{15pt}\noindent\textbf{Contrastive loss:} We apply all contrastive losses with a temperature hyperparameter (denoted by $Temp=0.25$), dividing each cosine similarity in Eq.\ref{eq_total_loss},\ref{eq_coarse_loss}. The value of $Temp$ was selected using a validation set (details about hyper-parameter search below).

\vspace{15pt}\noindent\textbf{Ground-truth regularization for training the set encoder $\ftheta$:}
For training the set encoder $\ftheta$, we use a regularization term that maximizes the cosine similarity of the predicted word embedding $\w^0_c$, with its ground truth word embedding $\vec{g}_c$, keeping $\w^0_c$ close to its ground truth value.
Namely, $\ell_{GT}(\w_c^0, \vec{g}_c) = -\cossim{\w_c^0}{\vec{g}_c}$, where $\vec{g}_c$ is the word embedding of the concept type  (e.g. the embedding of  ``dog''). If the concept type includes more than a single word, we take the first one.

\vspace{15pt}\noindent\textbf{Architecture:} When using CLIP, we always used  ViT-B/32 Vision Transformer.

\vspace{15pt}\noindent\textbf{Normalized embeddings:}
Wherever we use a textual or visual encoder output, we first normalize the embedding vector to unit norm. The embedding can be viewed as lying on a hypersphere.

\vspace{15pt}\noindent\textbf{Training the alignment matrix $\A$ with images and captions:}
In addition to updating the alignment matrix $\A$ during training of $\ftheta$ with text (Section 4.2), we also update $\A$ by mapping from captions to images.
Specifically, for every image $\img$ embedded with CLIP $v = \CLIPI(\img)$, we took the respective caption $S$ embedded with $u = \CLIPT(S)$, and trained $\A$ to project from the embedded captions to the embedded images by minimizing an $L_2$ loss:

\small
\begin{equation} \label{eq:alignment_loss}
    \ell\big(\img,S,\A\big) = || \CLIPI(\img) - \A  \CLIPT(S)  ||^2
\end{equation}

\vspace{15pt}\noindent\textbf{Using the alignment matrix $\A$ for the fine-tuning stage:} 
In practice, in the  fine tuning stage (\eqref{eq_coarse_loss}), we replace $\eta_c$ by $\A \cdot \eta_c$.

\vspace{15pt}\noindent\textbf{Training procedure for the set encoder $\ftheta$:}
We train $\ftheta$ in an alternating fashion. One batch with COCO images and one batch with augmented COCO captions.

\vspace{15pt}\noindent\textbf{Hyperparameters:}
Hyper parameters were tuned one at a time, on a validation set, to maximize the MRR metric in the retrieval task.

We train $\ftheta$ for 300 epochs. Batch size was set to $256$. We used the Adam \cite{kingma2014adam} optimizer with a learning rate of 0.0001 for both the cycle loss and the alignment loss (eq. \ref{eq:alignment_loss}). DeepSet's hidden dimension was set to 4096, and the dropout rate was set to 0.25. The contrastive loss temperature was set to $Temp=0.25$. To optimize the word embeddings, we used 30 epochs, with a learning rate of 0.01. The weight of the ground-truth regularization term $\lambda_{gt}$ was set to 512.

We used the following ranges to search for hyper parameters: (1) number of epochs $\in [100, 200, 300, 500, 1000]$ (2) batch size $\in [128, 256, 512]$ (3) learning rate  $\in [0.01, 0.001,  0.0001, 0.00001]$ (4) DeepSet's hidden dimension $\in [512, 1024, 2048, 4096]$ (5) dropout rate $\in [0.15, 0.25, 0.35, 0.5]$ (6) $Temp \in [0.15, 0.25, 0.35, 0.5]$ (7) number of fine-tuning personalization epochs $\in [10, 20, 30, 40, 50, 60]$ (8) fine-tuning learning rate $ \in [0.01, 0.001, 0.0001]$ (9) $\lambda_{gt} \in [1, 2, 4, 8, \dots 2048]$. %

\vspace{15pt}\noindent\textbf{Randomization:} \textit{Model:} For each of our $5$ repetitions we trained a new $\ftheta$ model. \textit{Few-shot training data:} When selecting a subset of few images from the few-shot training data, we made sure that the random seed (and subsets) are consistent between PALAVRA and the baselines (e.g. COLLIE, AvgIm, etc \dots ).

\vspace{15pt}\noindent\textbf{Training COLLIE and Adapter:}
To use multiple concepts that share the same concept type (category name), with the COLLIE and adapter baselines, we have assigned  a unique \concept{} phrase for concept. The phrase is composed of its class (e.g., skirt) and a unique ID number (e.g., ``skirt 241'').

\begin{figure}[h]
    \centering   
    \includegraphics[height=85pt, trim={0cm 0cm 0cm 0cm},clip]{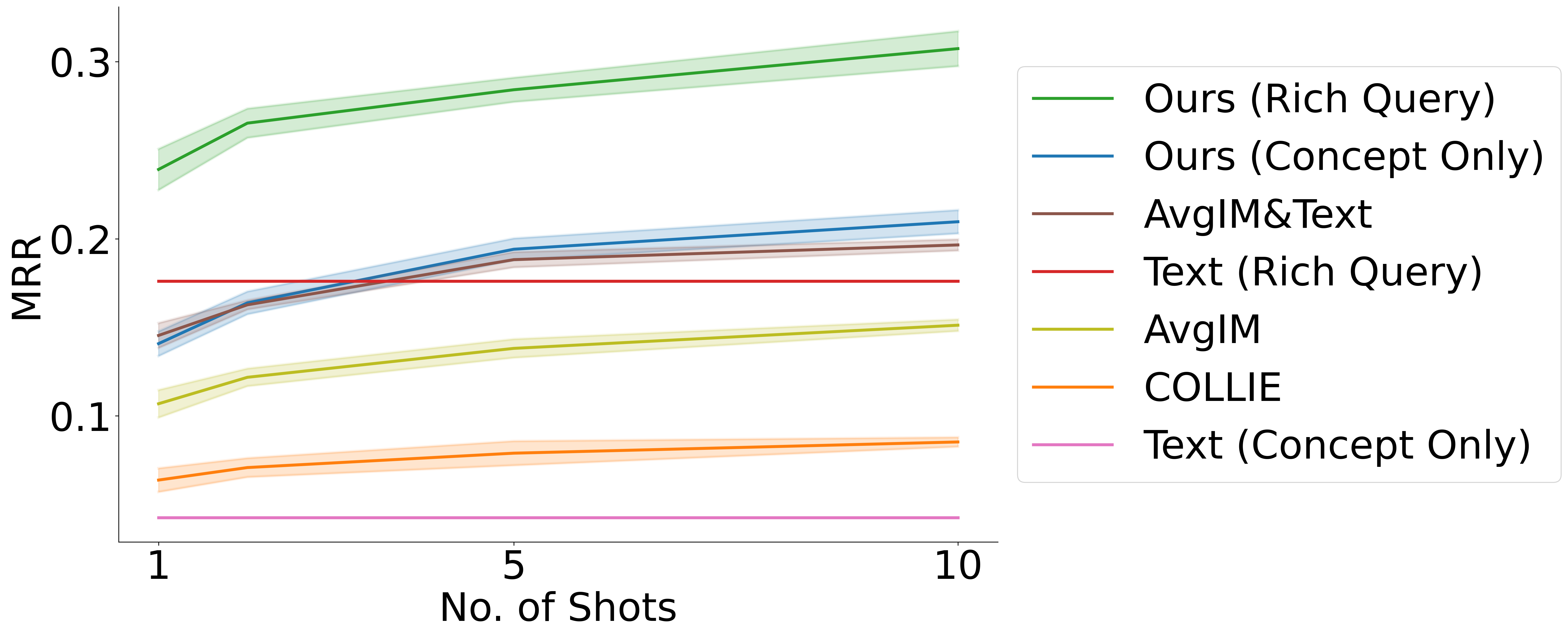} %
    \includegraphics[height=85pt, trim={0cm 0cm 0cm 0cm},clip]{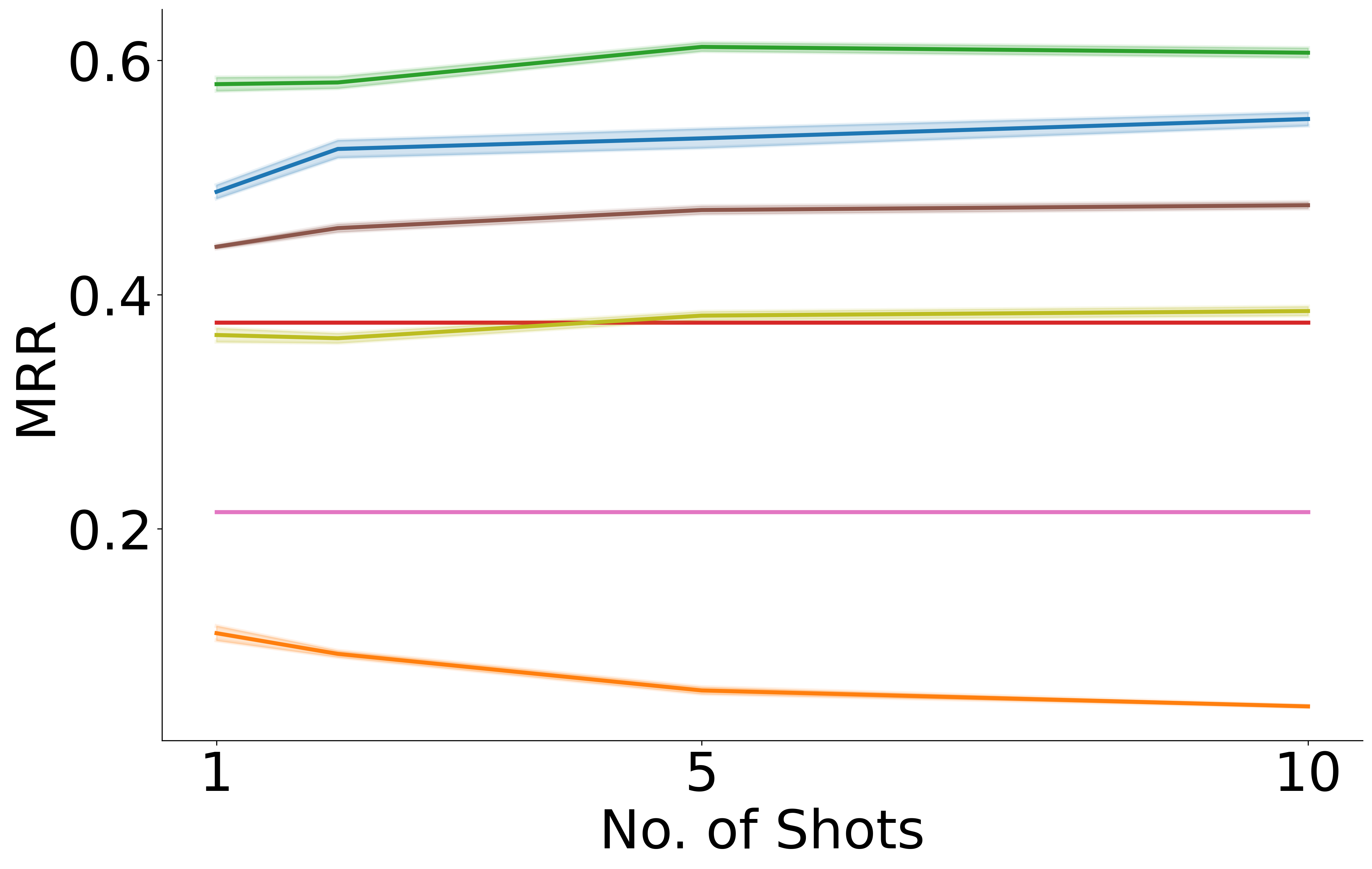} %
    \vspace{-5pt}
    \caption{ MRR for image retrieval on DeepFashion2 (left) and YTVOS (right) as function of the number of shots used to learn each personalized concept. DeepFashion2 performance improves as we increase the number of shots. YTVOS saturates early, probably because the training images of each concept have less variability, since they were all extracted from the same video.
    }

    \label{fig_set_size}
    \vspace{-5pt}
\end{figure}
\section{Additional Results}
\label{sec_additional_results}
\vspace{15pt}\noindent\textbf{Accuracy versus number of shots:}
\figref{fig_set_size} shows the performance of our model and the baselines as a function of the number of few-shot training samples used to learn each personalized concept. DeepFashion2 performance improves as the number of shots increases. YTVOS saturates early, probably because the training images of each concept have less variability since they were all extracted from the same video.

\vspace{15pt}\noindent\textbf{Short versus detailed captions:}
Table \ref{tab:long_captions} shows retrieval results on DeepFashion2 when using longer, detailed captions. As expected, all text-based methods demonstrate an increase in retrieval metrics across the board, indicating that they can successfully leverage additional information. Our method remains at the front even in this scenario, highlighting that the benefits of personalized concepts persist even when detailed descriptions are provided.

\setlength{\tabcolsep}{4pt}
\begin{table}[htbp]
\centering
    \begin{tabular}{llll}
    \toprule
    {} &               MRR &          Recall@5 &         Recall@10 \\
    \midrule
    PALAVRA (Ours)      &  \textbf{33.8 $\pm$ 0.5\% } &  \textbf{47.5 $\pm$ 0.9\% } &  \textbf{61.9 $\pm$ 0.9\%  } \\
    PALAVRA w.o. tuning &  27.8 $\pm$ 0.3\% &  36.4 $\pm$ 0.6\% &  48.4 $\pm$ 0.6\% \\
    AvgIM+Text          &  20.9 $\pm$ 0.6\% &  29.0 $\pm$ 0.7\% &  38.4 $\pm$ 0.6\% \\
    Text (CLIP)         &  24.3 $\pm$ 0.0\% &  31.7 $\pm$ 0.0\% &  43.4 $\pm$ 0.0\% \\
    \bottomrule
    \end{tabular}
    \vspace{2pt}
    \caption{Retrieval results using detailed captions. As expected, all compared methods show improved performance when provided with extra textual information. Notably, our method maintains the advantage even in such a scenario, showing that it can yield an increase in performance even when the concepts are described in detail.}
    \label{tab:long_captions}
\end{table}

In \secref{sec_short_vs_detailed_AMT} below we explain the data collection procedure of the ``detailed'' and ``short'' queries.

\vspace{15pt}\noindent\textbf{COLLIE sensitivity to prompt:}

In \secref{sec_retrieval_results} we demonstrated that COLLIE performance degrades when using rich textual queries. Here we describe results showing that COLLIE is even sensitive to much simpler queries. Namely, template queries that only add a \textit{prefix prompt}.

When the text query includes only the \concept{} tag, as in COLLIE's training procedure, COLLIE achieves a 13.4\% average MRR score on DeepFashion2 retrieval test set. When the query text is a sentence with a prefix, its score drops sharply. For example, the query ``This is a photo of a \concept{}" results in an MRR score of 5.3\%,  the query ``This looks like a \concept{}" score is 4.6\%, and the query ``In this image, there is a \concept{}" yields 3.3\%.

\begin{figure}[htbp]
    \centering
    {\scriptsize
    \begin{tabular}{cccc|cc}
        
        \quad\quad~Ours & \quad\quad Ground Truth & \quad\quad~Ours & \quad\quad Ground Truth & \quad\quad~Ours & \quad\quad Ground Truth \\
        
        \multicolumn{2}{c}{\includegraphics[width=0.315\textwidth, trim={0cm 0cm 0cm 0cm},clip]{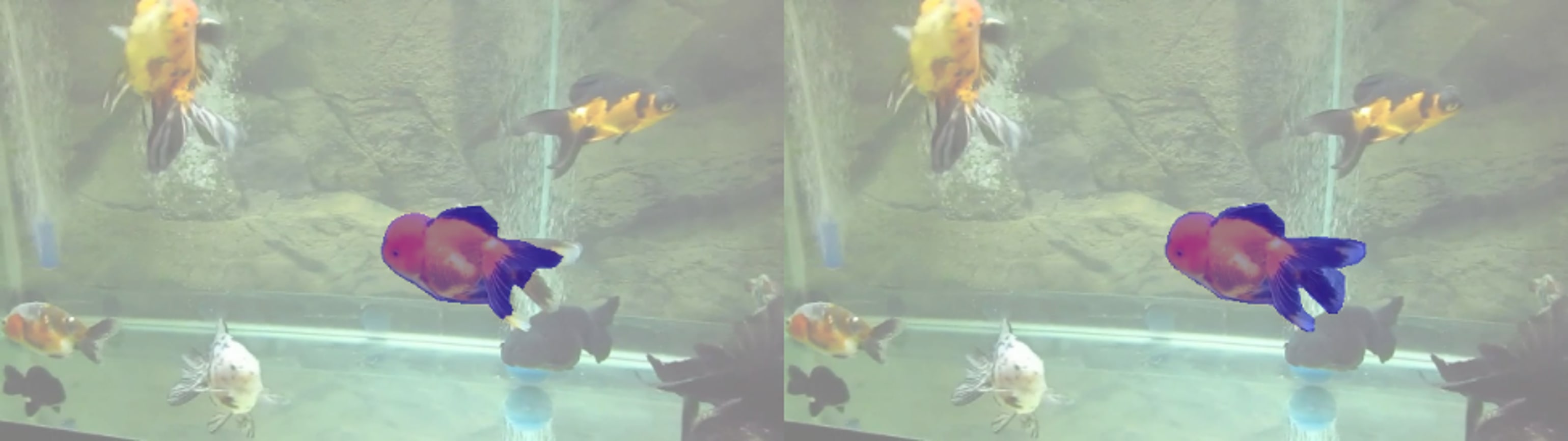}} &
        \multicolumn{2}{c|}{\includegraphics[width=0.315\textwidth, trim={0cm 0cm 0cm 0cm},clip]{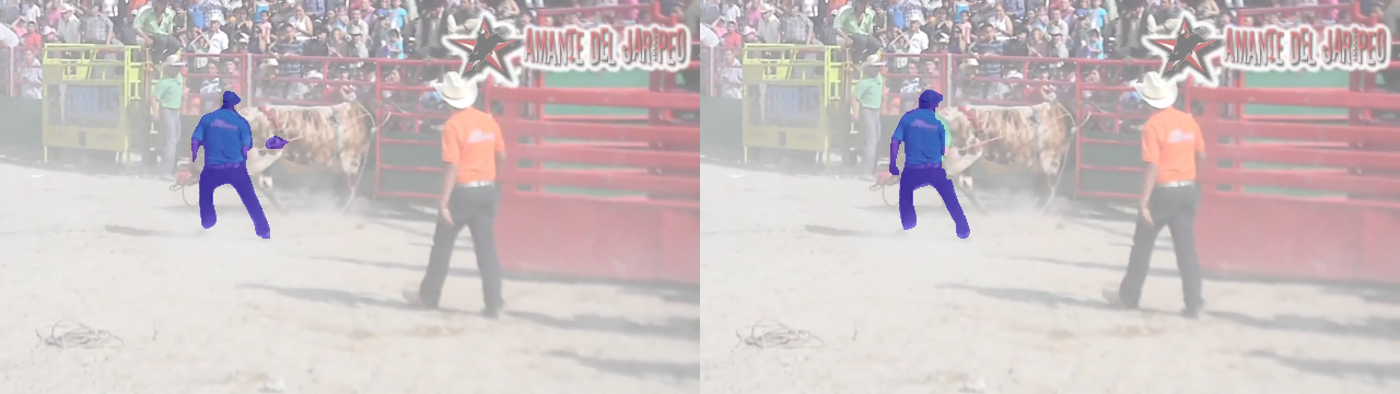}} &
        \multicolumn{2}{c}{\includegraphics[width=0.315\textwidth, trim={0cm 0cm 0cm 0cm},clip]{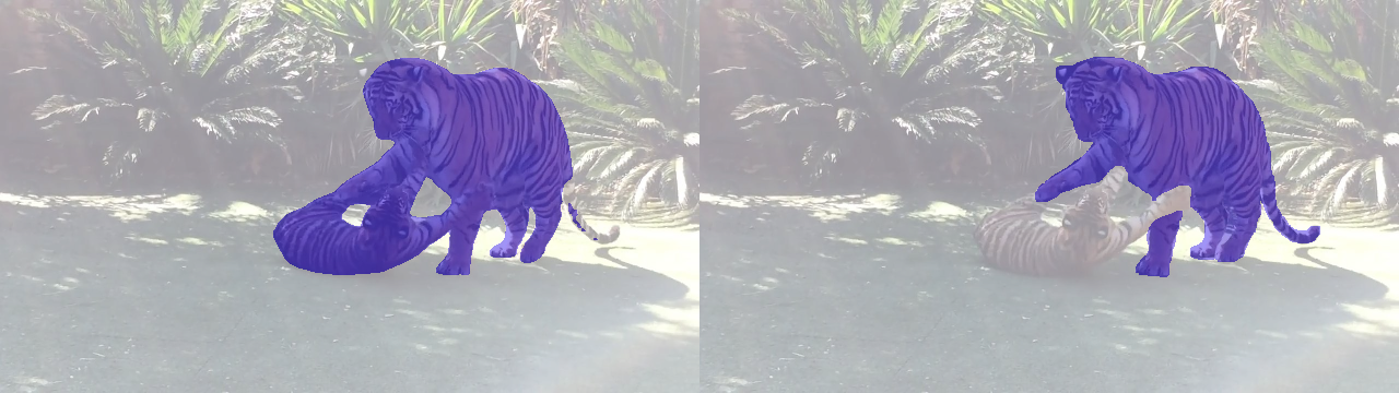}} \\
        \multicolumn{2}{c}{\tiny A bright orange [CONCEPT]} &
        \multicolumn{2}{c|}{\tiny A [CONCEPT] } &
        \multicolumn{2}{c}{\tiny A [CONCEPT] standing} \\
        \multicolumn{2}{c}{\tiny with its full black dorsal fin and} &
        \multicolumn{2}{c|}{\tiny wearing a green shirt} &
        \multicolumn{2}{c}{\tiny above another tiger} \\
        \multicolumn{2}{c}{\tiny  black tail with white tips visible} &
        \multicolumn{2}{c|}{\tiny and black jeans} &
        \multicolumn{2}{c}{\tiny } \\
        \multicolumn{2}{c}{\includegraphics[width=0.315\textwidth, trim={0cm 0cm 0cm 0cm},clip]{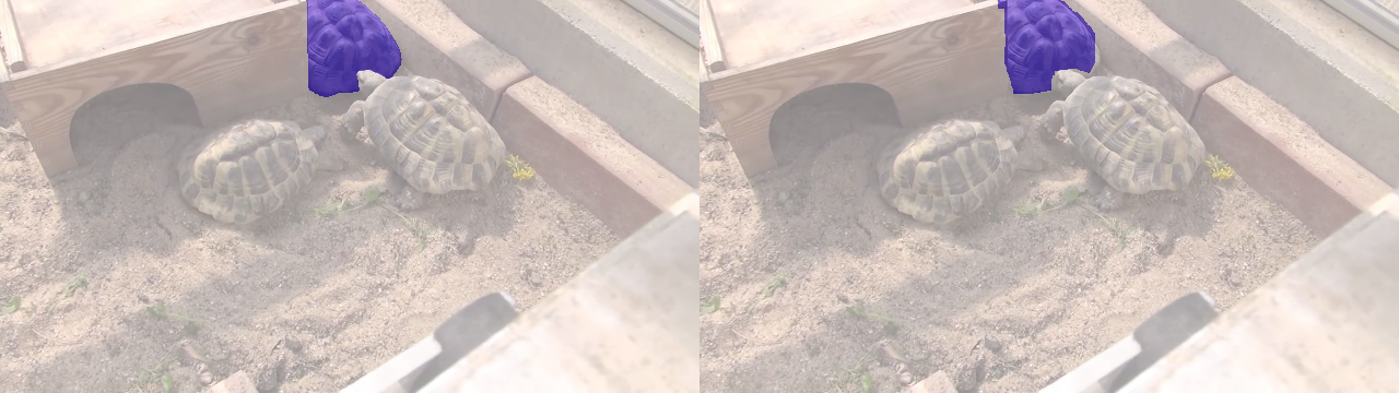}} &
        \multicolumn{2}{c|}{\includegraphics[width=0.315\textwidth, trim={0cm 0cm 0cm 0cm},clip]{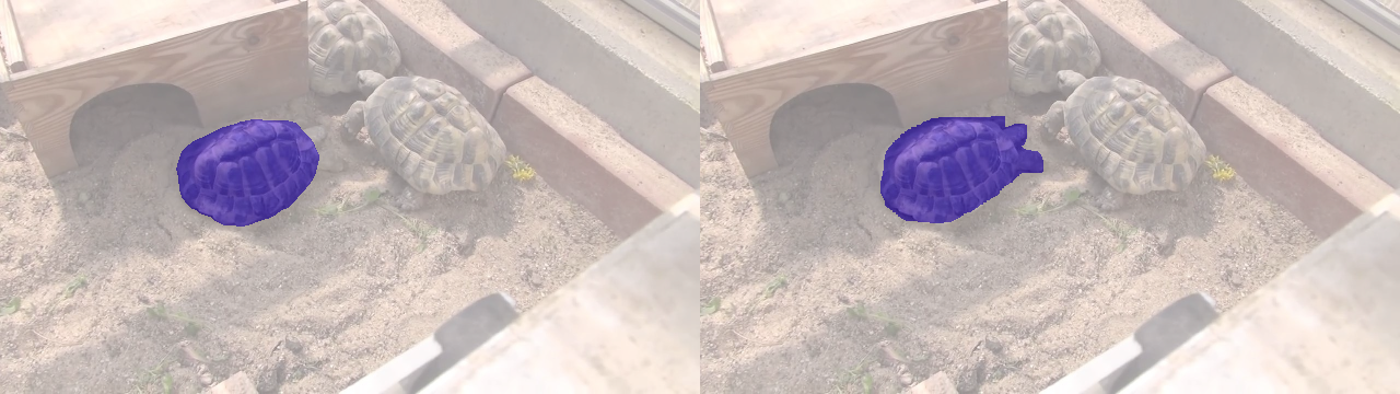}} &
        \multicolumn{2}{c}{\includegraphics[width=0.315\textwidth, trim={0cm 0cm 0cm 0cm},clip]{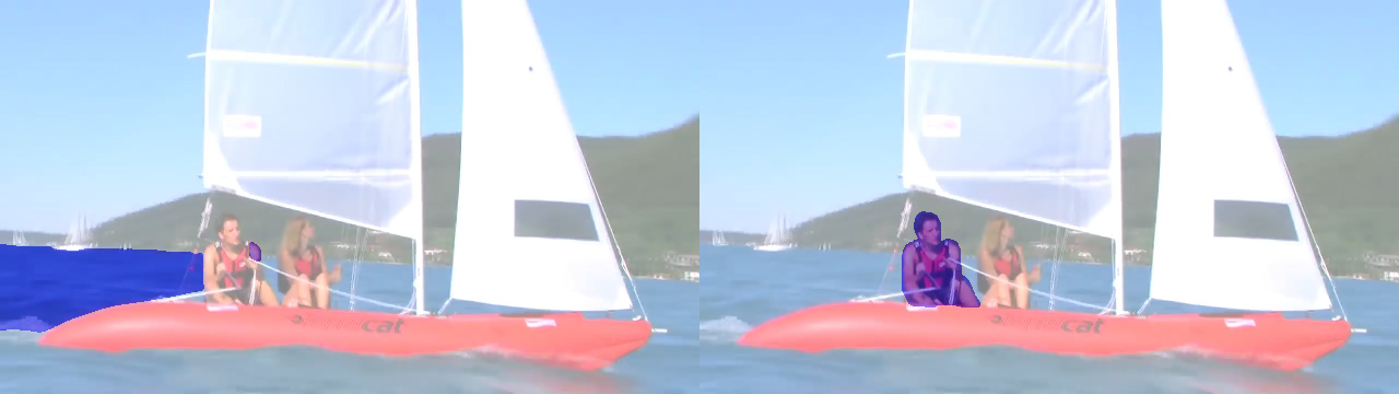}} \\
        \multicolumn{2}{c}{\tiny A [CONCEPT]}        &
        \multicolumn{2}{c|}{\tiny A [CONCEPT]} &
        \multicolumn{2}{c}{\tiny [CONCEPT] sits at} \\
        \multicolumn{2}{c}{\tiny wedged between brick and wood}        &
        \multicolumn{2}{c|}{\tiny standing next to a doorway} &
        \multicolumn{2}{c}{\tiny the back end of the sailboat} \\  & & & & &\\
        \multicolumn{4}{c|}{Wins}        &
        \multicolumn{2}{c}{Losses} \\
    \end{tabular}
    }
    \caption{Qualitative examples of PALAVRA used for semantic segmentation. \textbf{Left and middle:} successful segmentations, where the correct specific object is identified and extracted from the image despite similar distractors (other turtles). \textbf{Right:} Typical failure cases: distractors are segmented along with the personalized object (top), or the textual descriptions draw CLIP's attention away from the main object (bottom).}
    \label{fig:seg_qualitative}
\end{figure}

\vspace{15pt}\noindent\textbf{Qualitative results for semantic segmentation:}
In Fig. \ref{fig:seg_qualitative} we show curated qualitative segmentation results. 
We observe that our model can successfully segment the correct object instance even in scenarios with visually similar distractors. On the other hand, our model can sometimes fail to distinguish between multiple relevant candidates, or segment other objects which exist near the target. However, this last limitation may be an artifact of the underlying segmentation method. %

\section{Personalization of Other Vision \& Language}
\label{approch_extension}
Vision \& Language models other than CLIP may also benefit from an extended vocabulary of personalized concepts. It is likely that a similar approach to ours can still be applied. For example, in models like M-DETR \cite{kamath2021mdetr}, $f_\theta$ could map from the CNN output space to the \textit{input} space of RoBERTa. The alignment matrix $\A$ can close the cycle, mapping from the \textit{output} of RoBERTa to the CNN output.

\section{Segmentation Details and Analysis}

\subsection{Baselines and hyper parameters}

Our segmentation experiments use the framework of Zabari~\etal\cite{zabari2021semantic}. The method leverages transformer interpretability methods to identify image regions that relate to a given textual prompt. In this process, the text-encoding branch is only used to supply an embedding vector which is matched with the image branch. As such, the embedding vector can be easily replaced with another vector from any source. We leverage this property for all of our baselines.

When conducting an image-based search (AvgIM), we replace the embedding vector with the normalized average embedding of a small set of images depicting the target object. For the AvgIM\&Text baseline, we further average this image embedding with the text embedding of the query text.

To compare with COLLIE, we generate the embeddings using their adapter setup. For our method, we utilize the original CLIP text encoder but substitute our learned input word embeddings for the concept token.

Hyper parameters were tuned on the validation set and kept fixed for all methods. We use a resizing factor of $0.5$ and generate $3$ `clicks' from the relevancy maps for the single image segmentation method. All other parameters were unchanged from the baseline implementation of Zabari~\etal\cite{zabari2021semantic}.

\subsection{Rich queries versus concept-only queries}
\label{app:rich_mask_vis}
In the main manuscript, we noted that surprisingly the segmentation model performed better when provided with a text target of the form ``A photo of a \concept{}" (i.e. a ``Concept-only" query) than when provided with a rich textual caption.

To investigate this behavior, we turn to an analysis of the local relevance maps, which are used to guide the segmentation. Our investigation reveals that often, when the rich query describes other objects within the image, CLIP's attention drifts towards those objects. That is, CLIP struggles with leveraging relational information in the text and instead splits its focus between several objects mentioned in the rich query. Figure \ref{fig:relevancy_drift} provides a qualitative visualization of this effect.

\begin{figure}[htbp]
    \centering
    \includegraphics[width=0.7\textwidth]{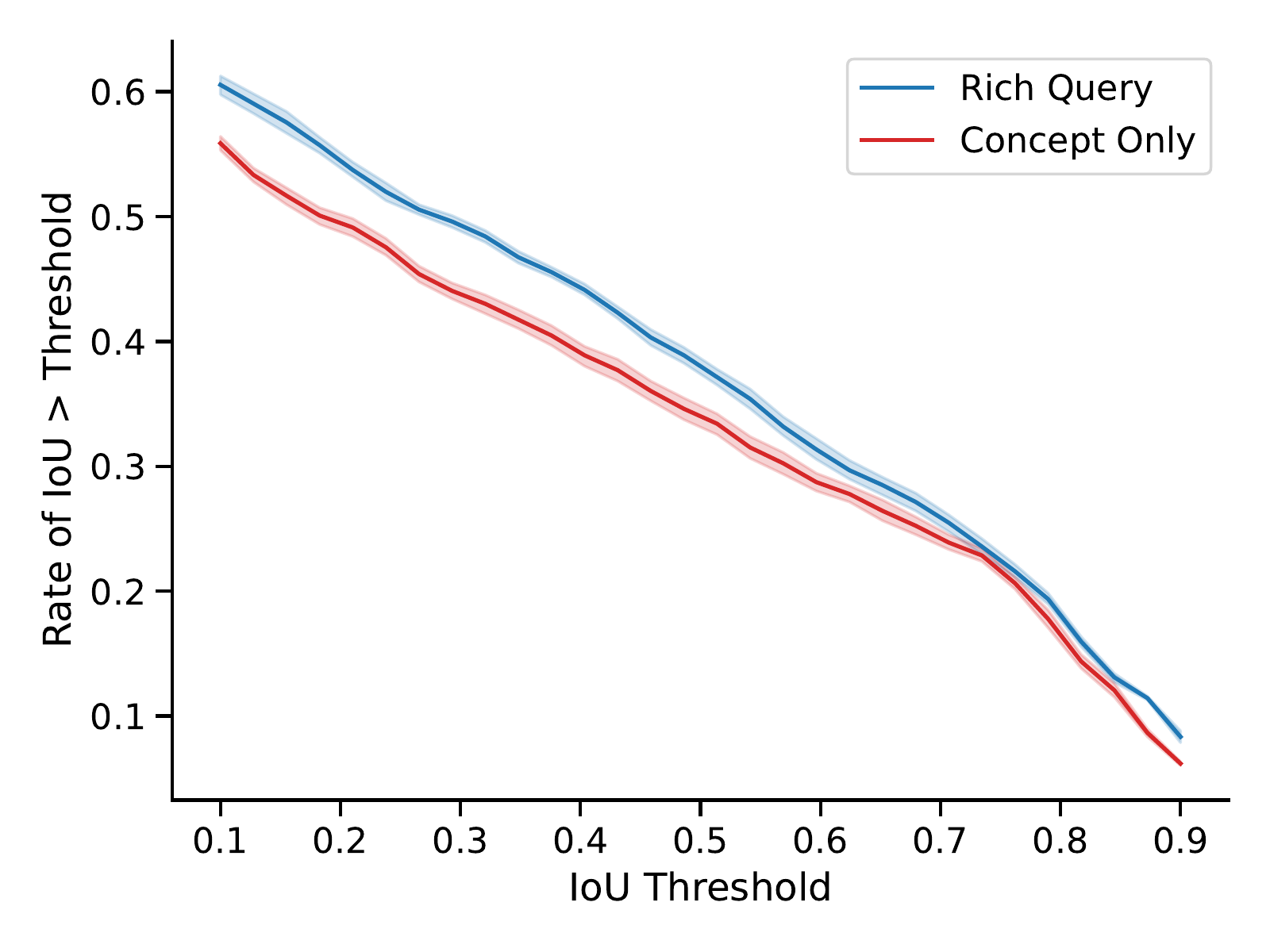}
    \caption{Rich queries
outperform concept-only
queries when context objects are not valid segmentation candidates.}
    \label{fig:seg_mask_context}
\end{figure}

\edit{
To \textit{quantitatively test this hypothesis}, we re-ran segmentation, this time masking out relevancy scores of the background, except for objects which are also valid retrieval candidates. Now, context objects were no longer valid candidates. Indeed, we found that with this manipulation, rich queries do outperform concept-only queries, as in retrieval (Fig. \ref{fig:seg_mask_context}). %
}

We conclude that a good caption for CLIP-guided personalized segmentation should describe the object or its immediate vicinity, and not its relation to other objects.

\begin{figure}[htbp]
    \centering
    \includegraphics[width=0.7\textwidth]{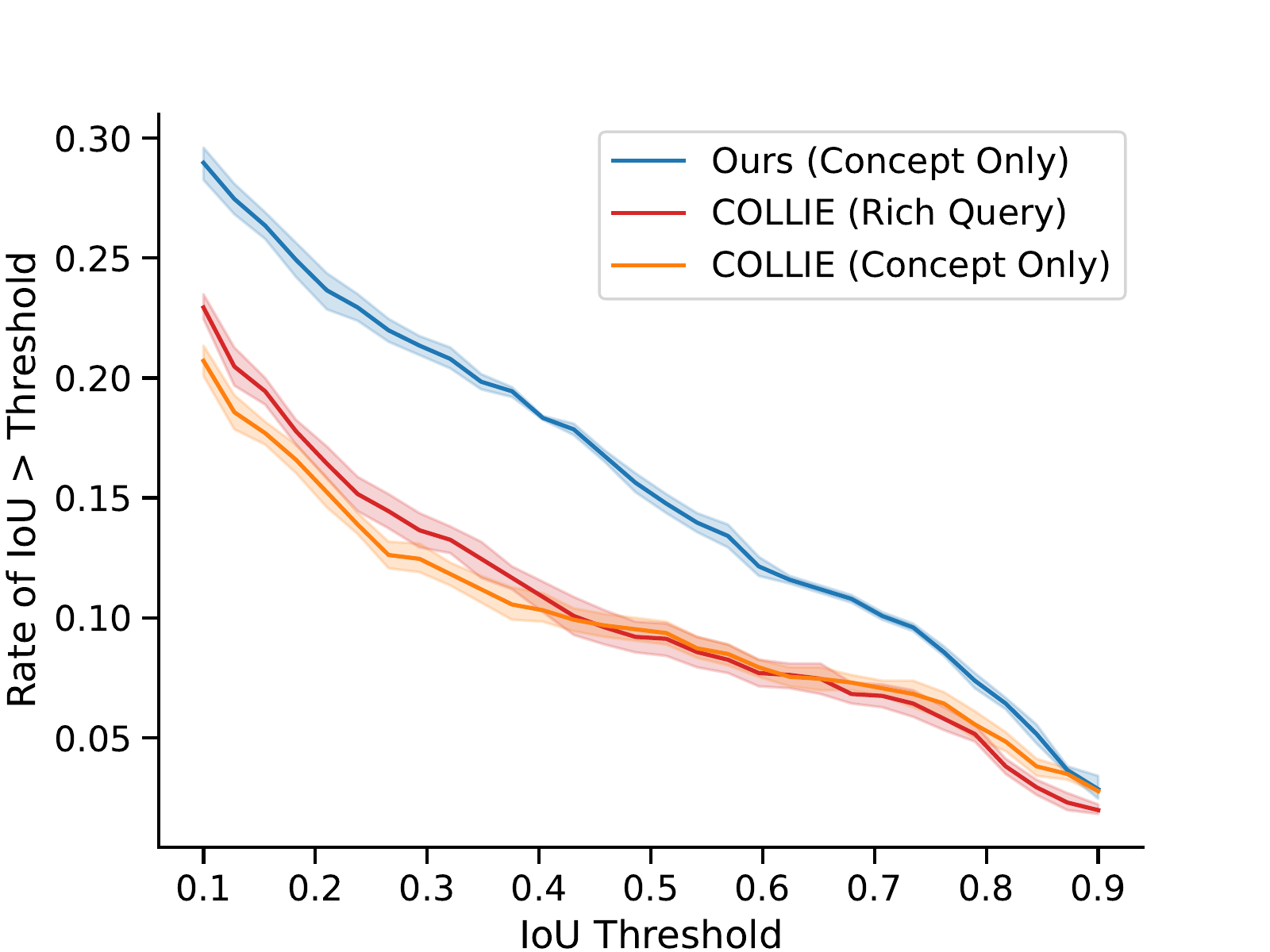}
    \caption{COLLIE performance when supplied with rich queries and with  concept-only queries. The performance of COLLIE does not depend on the rich query, indicating that the additional information is largely ignored in the case of segmentation.}
    \label{fig:collie_sup}
\end{figure}

Last, we further investigated whether COLLIE demonstrates a similar sensitivity to rich queries. COLLIE's segmentation performance in the two scenarios is shown in \figref{fig:collie_sup}. We observe that, in contrast to our own approach and the baseline CLIP, COLLIE's performance on the segmentation task does not appear to be sensitive to the level of detail in the query.

\subsection{Qualitative Analysis}

\begin{figure}[htbp]
    \centering
    \includegraphics[width=1.02\textwidth,  trim={0cm 3cm 0cm 0cm},clip]{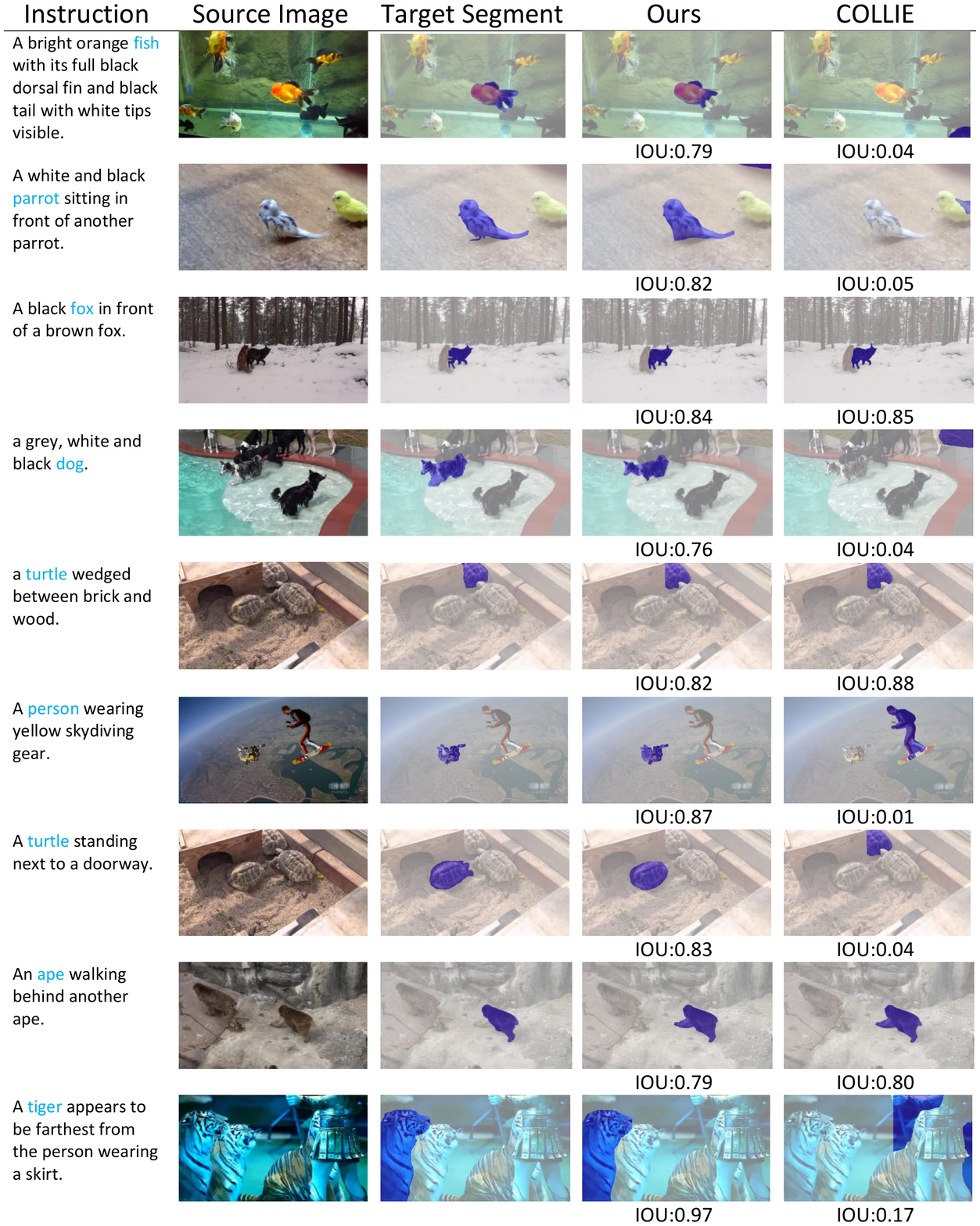}
    \caption{Examples of successful segmentation. For visualization purposes, we replaced the \concept{} tag by the name of its concept type, and highlighted it in cyan. \label{app:wins}}
\end{figure}

\begin{figure}[htbp]
    \centering
    \includegraphics[width=1.05\textwidth,trim={0cm 5cm 0cm 0cm},clip]{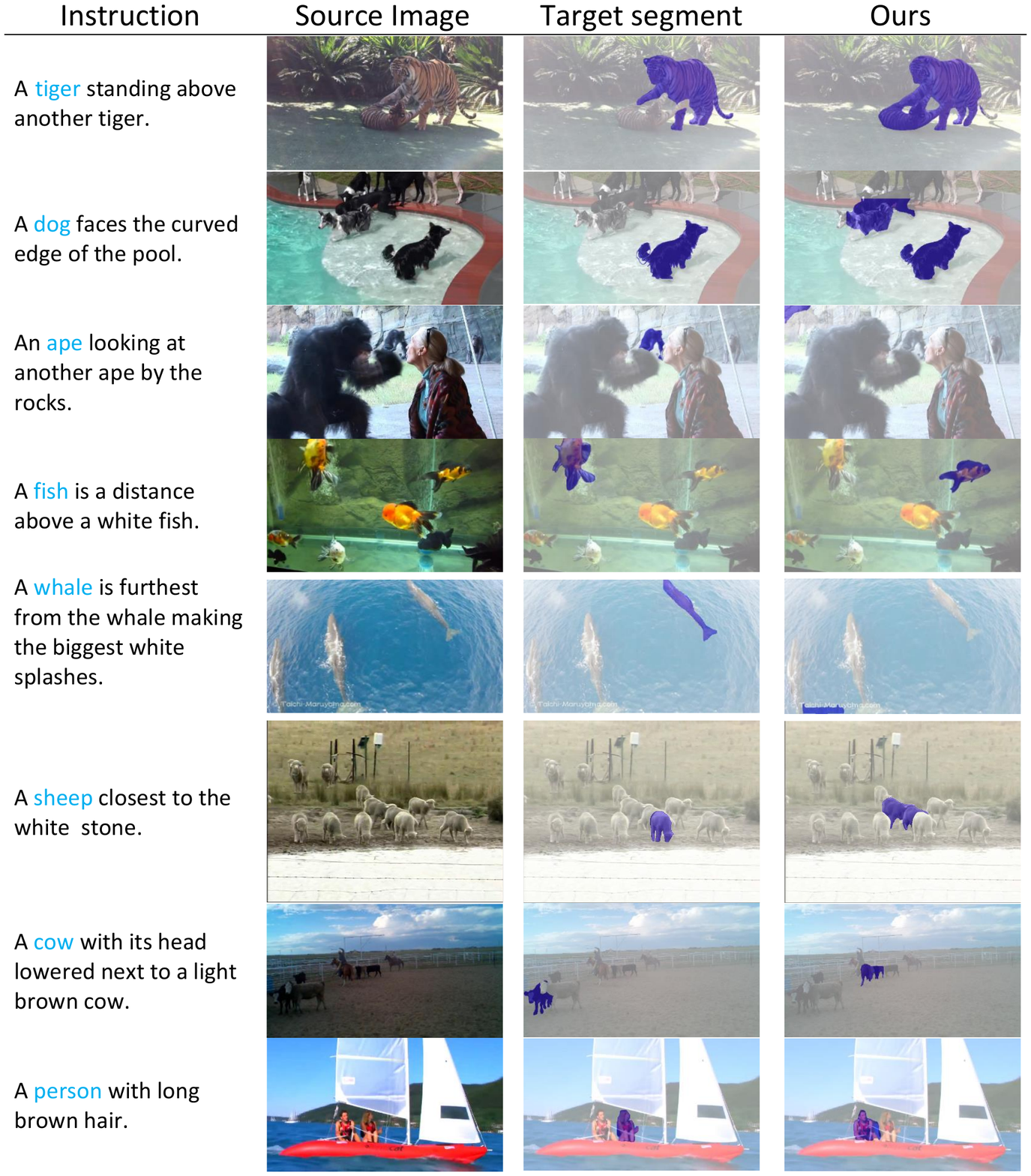}
    \caption{Examples of segmentation failures. For visualization purposes, we replaced the \concept{} tag by the name of its concept type, and highlighted it in cyan. \label{app:losses}}
\end{figure}

\begin{figure}[htbp]
    \centering
    {\scriptsize
    \begin{tabular}{ccc}
        Image & Concept-Only & Rich Query \\
        \includegraphics[width=0.315\textwidth]{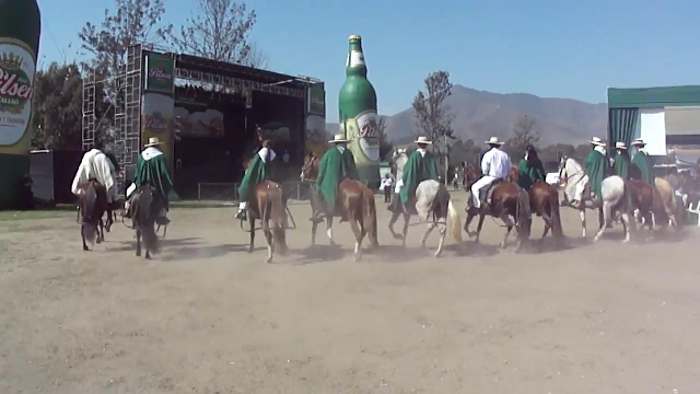} &
        \includegraphics[width=0.315\textwidth]{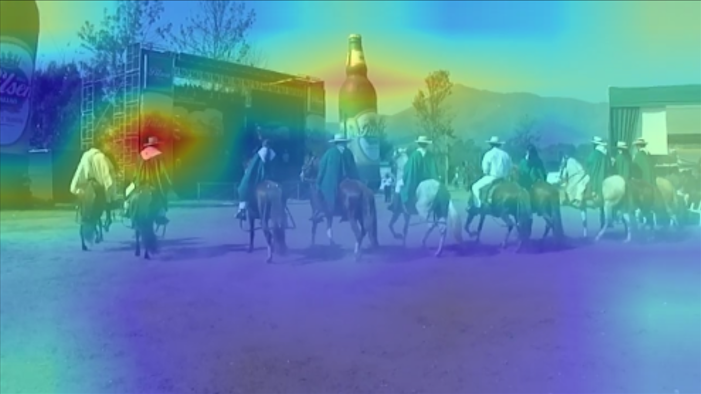} &
        \includegraphics[width=0.315\textwidth]{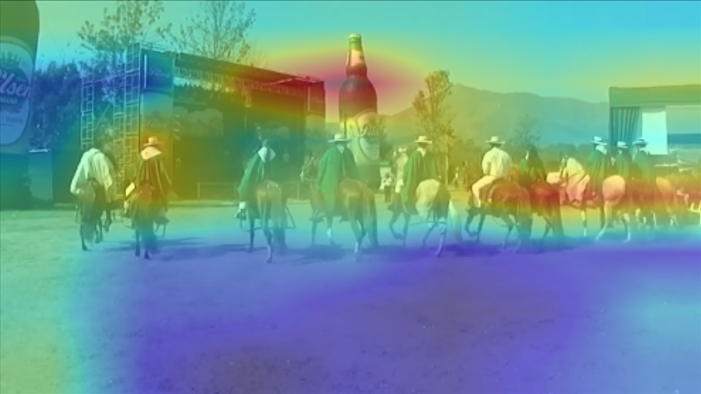} \\
        \multicolumn{3}{c}{``\concept{} is on the head of the rider in a green poncho on a medium-brown horse"} \\
        \includegraphics[width=0.315\textwidth]{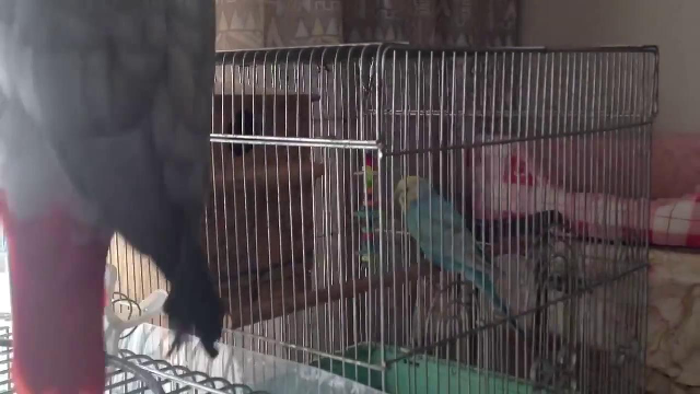} &
        \includegraphics[width=0.315\textwidth]{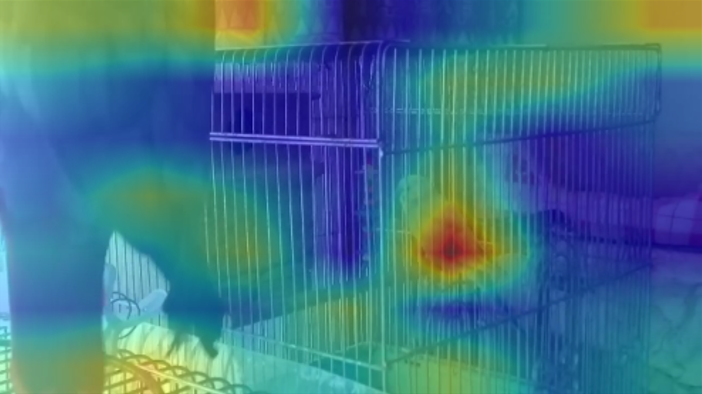} &
        \includegraphics[width=0.315\textwidth]{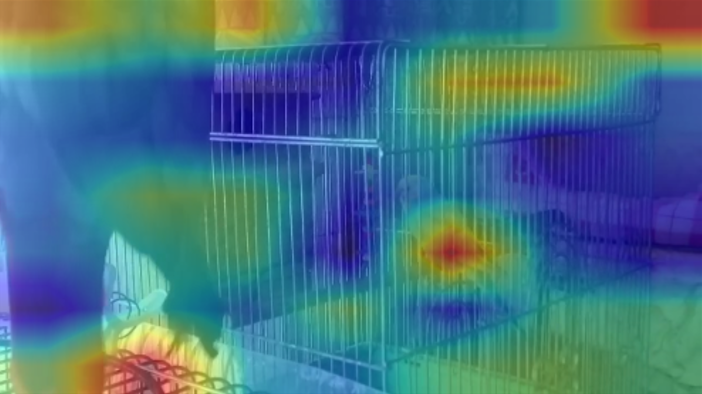} \\
        \multicolumn{3}{c}{``A \concept{} perched inside a cage."} \\

    \end{tabular}
    }
    \vspace{-5pt}
    \caption{Qualitative examples of `attention drift' when using rich queries. When the descriptor mentions other objects, CLIP's attention visually drifts away from the target concept and towards other objects described in the query. For example, in the top row, focus moves from the hat and towards the brown horses. On the bottom row, focus moved away from the parrot and towards the empty cage at the bottom of the frame.}
    \label{fig:relevancy_drift}
\end{figure}

In this section, we provide an additional qualitative analysis of segmentation using PALAVRA. We compare our results with the recent baseline COLLIE. Figure \ref{app:wins} shows examples of successful segmentation, and Figure \ref{app:losses} shows some failure modes.

\section{Evaluation datasets}
\label{sec_datasets_suppl}
We provide details for creating our two new benchmark datasets, based on DeepFashion and YTVOS.

\subsection{DeepFashion2}
\label{sec_deepfashion_suppl}

To ensure that DeepFashion2 benchmark items are included in a rich visual context, items were included in the dataset if they obey the following criteria: (1) Have at least 5 images with a proper scale (zoom). Specifically, the item covers no more than $50\%$ of the image. (2) There are at least 15 images of the same item in total. The set yielded $\tildeapprox 1700$ images and 100 unique fashion items (concepts) which met these criteria. Each unique fashion item was assigned a unique \concept{} tag.

Next, we explain how we annotated a subset of this data with textual descriptions, and how we selected the evaluation set.

We manually curated a subset of 652 images (out of 1700) that contain a person wearing a fashion item and at least one additional object for a context.
We did not consider mobile phones or mirrors as valid context, as these objects are abundant in the dataset. For each image, we collected a textual description that refers to each fashion item. For instance, \textit{The \concept{} is facing a glass store display."}.

To provide diverse captions, we instruct the raters to \textit{avoid} trivial captions such as \textit{``a \concept{} in front of a mirror''}. We also instructed them to avoid describing the item itself, because the same item appears in several evaluation images, and we wished to have a textual query which is specific to one single image.

We randomly sampled an evaluation set (out of the 652 images), by sampling 5 images per concept, or less, if not available.
This results in 450 evaluation images and 1250 images for training. Finally, we made a concept-based split by randomly splitting the dataset to 50 validation concepts and 50 test concepts.

\begin{figure}[htbp]
\label{AMT_describe_person}
    \centering
    \includegraphics[width=0.93\textwidth, trim={0.1cm 4cm 10.7cm 0cm},clip]{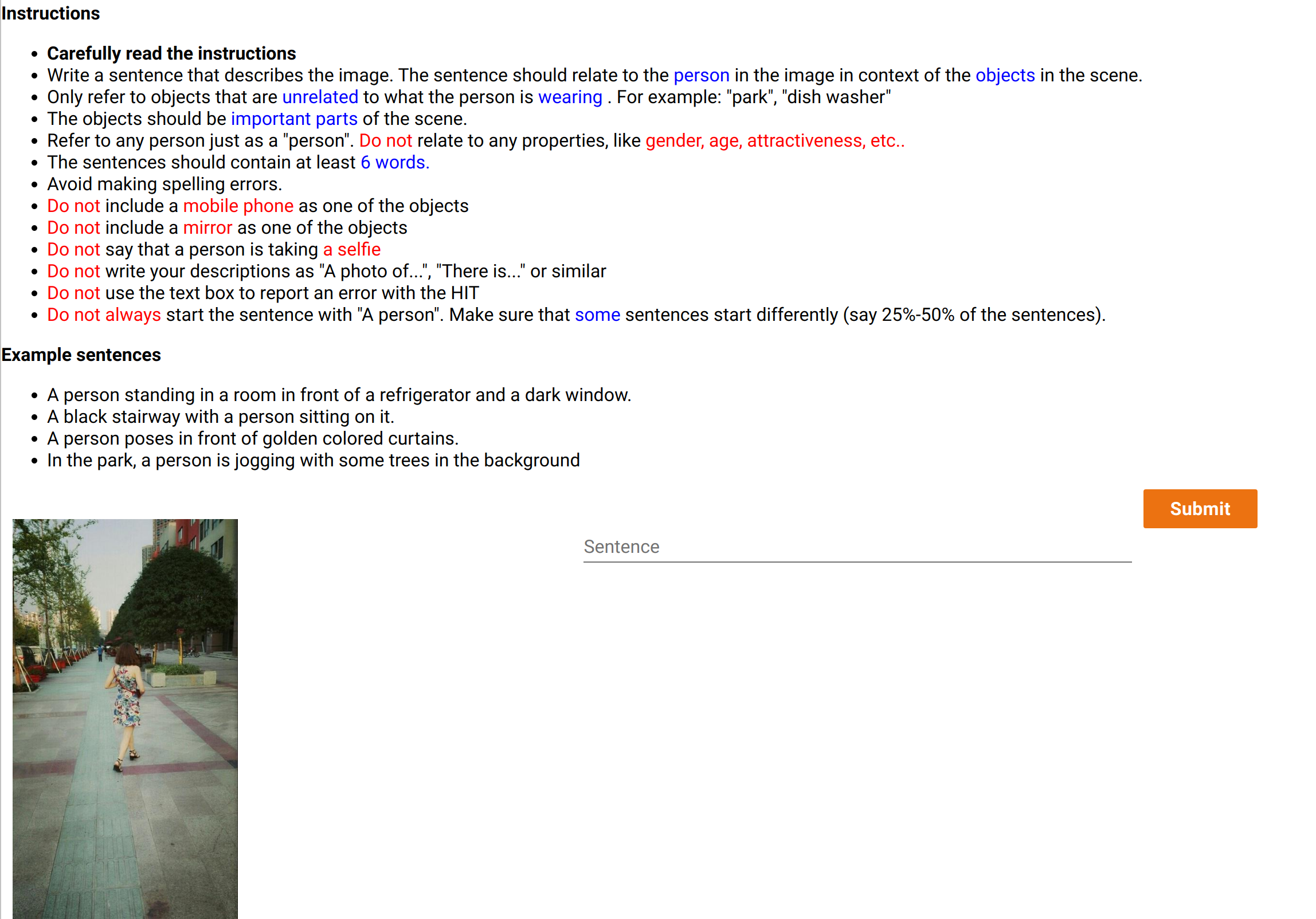} %
    \hspace{-10pt}
    \vspace{-10pt}
    \caption{
Instructions for collecting textual descriptions for images of the DeepFashion2 benchmark.
    }
    \label{fig_AMT_describe_person}
    \vspace{-10pt}
\end{figure}

\begin{figure}[htbp]
    \centering
    \includegraphics[width=0.93\textwidth, trim={0.1cm 0cm 4cm 0cm},clip]{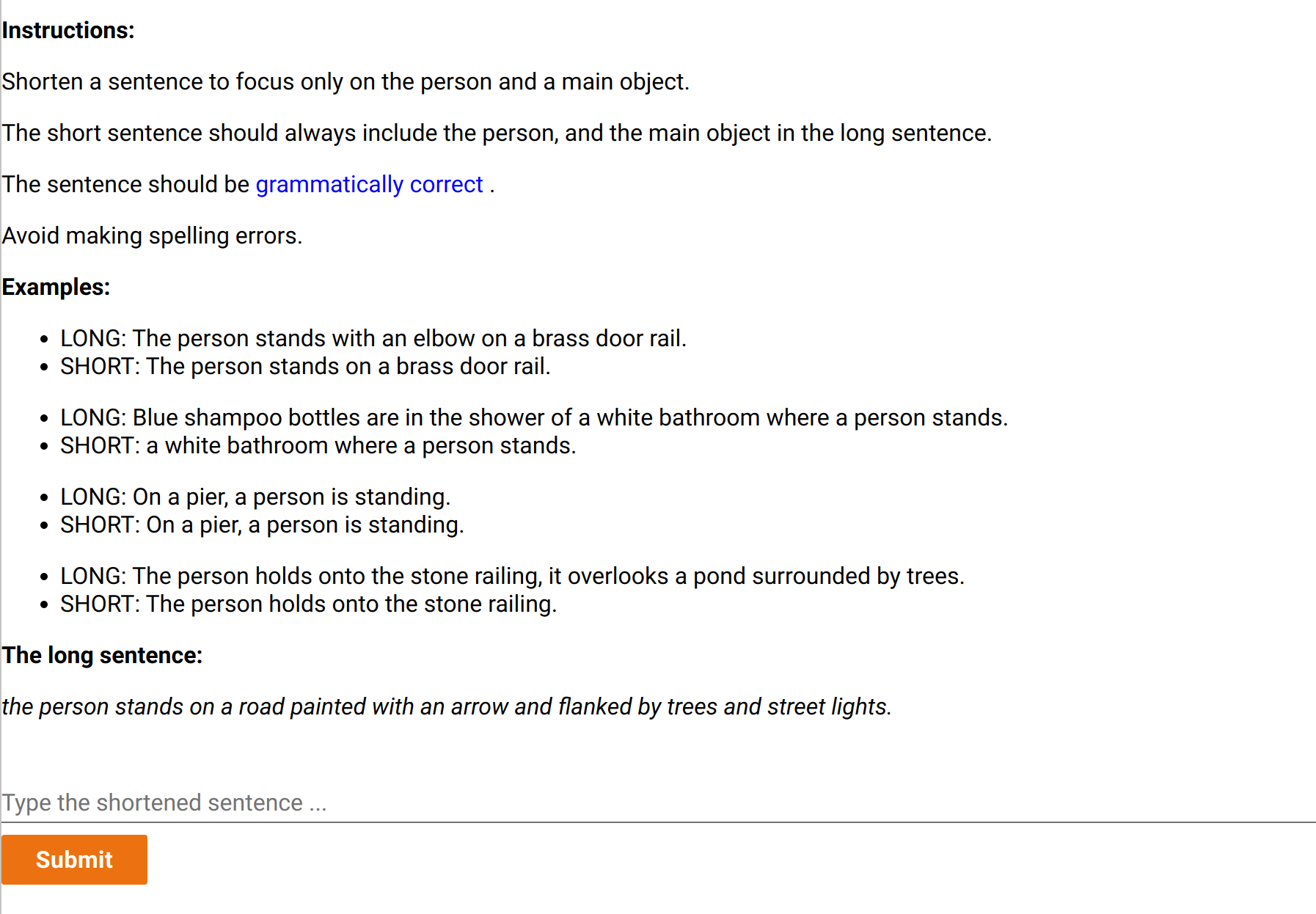} %
    \hspace{-10pt}
    \vspace{-10pt}
    \caption{
Instructions for summarizing textual descriptions for images of the DeepFashion2 benchmark.
}
    \label{fig_AMT_shorten_caption}
    \vspace{-5pt}
\end{figure}

\vspace{5pt}\noindent\textbf{Annotations for DeepFashion2 with Amazon Mechanical Turk}

To simplify the instructions for collecting textual annotations, we used the fact that every fashion item is worn by a person.
When describing the images, we simply asked the raters to relate to the person in the image in context of the objects in the scene, and in a post-processing step, we replace every mention of the word ``person'' by the ``\concept{}'' token. Additional instructions were inspired by the instructions provided for collecting captions for the COCO dataset (See appendix of \cite{lin2014microsoft}).

Finally, to maintain the quality of the textual descriptions, we only worked with the raters after they passed our qualification test, making sure that they followed the instructions when describing 5-10 images. In addition, we only worked with raters with AMT ``masters'' qualification, demonstrating a high degree of approval rate in a wide range of tasks. We paid the raters $0.2\$$ for annotating each image.

\figref{fig_AMT_describe_person} provides an example of the data collection API for textual annotation of images for the DeepFashion2 benchmark.

\subsection{Summarizing textual annotations with AMT}
\label{sec_short_vs_detailed_AMT}

As explained in \secref{sec_additional_results}, for DeepFashion2 we created two types of captions for each image, in order to quantify the effect of caption length. We expected that image retrieval with short textual queries will pose a greater challenge, because they contain less information about the target image, leading to queries that are more ambiguous.

To create the set of ``short'' text queries, we took the set of image captions described in \secref{sec_deepfashion_suppl}, which we now denote as ``detailed'' captions, and asked the AMT raters to summarize each caption. Given a detailed caption, their goal was to describe the concept in the context of a \textit{single} object in the scene.
An example of a caption and its summarized version is:
``\textit{White cabinets, some with open drawers, are alongside and behind the \concept{}.}'' was summarized to ``\textit{White cabinets are behind the \concept{}.}''

\figref{fig_AMT_shorten_caption} provides an example of the data collection API to summarize textual descriptions.

Similarly to the previous section, to maintain the quality of the textual descriptions, we only worked with raters after they passed our qualification test and have AMT ``masters'' qualification.
We paid the raters $0.1\$$ for summarizing each caption.

Finally, for most of the DeepFashion2 experiments throughout the paper, we used the more challenging ``short'' queries. In \secref{sec_additional_results} we describe the evaluation results with the ``detailed'' queries.

\subsection{Youtube-VOS}
\label{app:ytvos_details}
\vspace{5pt}\noindent\textbf{Overview}

We created an image segmentation benchmark of personalized visual concepts given a textual query using Youtube-VOS (YTVOS)~\cite{xu2018youtube}. YTVOS is a dataset for instance segmentation in video, which includes 4000+ videos, 90+ categories, and 7800+ unique object instances. The original videos were $3-6$ second long with a 30 FPS frame rate. The dataset contains a subset of the frames, sampled at rate of 6 FPS. To transform the dataset into an image personalization benchmark, we take the last frame of each video (scene) for evaluation and the object instances that appear in it as target concepts. Earlier frames that contain that object are used as candidate frames for few-shot training. See examples in Figures \ref{fig_dataset_examples}(left), \ref{fig:seg_qualitative}, \ref{app:wins}, \ref{app:losses} and \ref{fig:relevancy_drift}.

For building the concept set, we consider each object instance (e.g. each animal in the frame) as a unique personalized concept. We chose training samples such that their object instantiations are not trivially solved by simple visual template matching with the last (evaluation) frame. To that end, we use the following criteria:  For each object instance, we consider all the previous video frames that contain it. We keep only the frames where: (1) the object's segmentation mask has a zero intersection-over-union (IOU) score when compared with its mask at the last frame (i.e. the evaluation target) and (2) the center of the mask moved at least 150 pixels when compared to the final frame. We discard any object instance that does not have at least 4 training examples left at the end of this filtering process. Finally, we take a box crop of the images around the selected masks and use them as training examples.

\vspace{5pt}\noindent\textbf{Annotation with AMT}

We annotated the instances in the evaluation frame with captions using AMT. We instructed the AMT workers to concisely describe what makes a specific entity distinct, compared with similar entities in the image, and, if possible, preferring descriptions that relate to one object that is nearby. %

Finally, similar to the previous sections, to maintain the quality of the textual descriptions, we only worked with raters after they passed our qualification test and have AMT ``masters'' qualification.
We paid the raters $0.3\$$ for every textual description.

\begin{figure}[h]
    \centering
    \includegraphics[width=0.93\textwidth, trim={0.cm 0cm 2.5cm 0cm},clip]{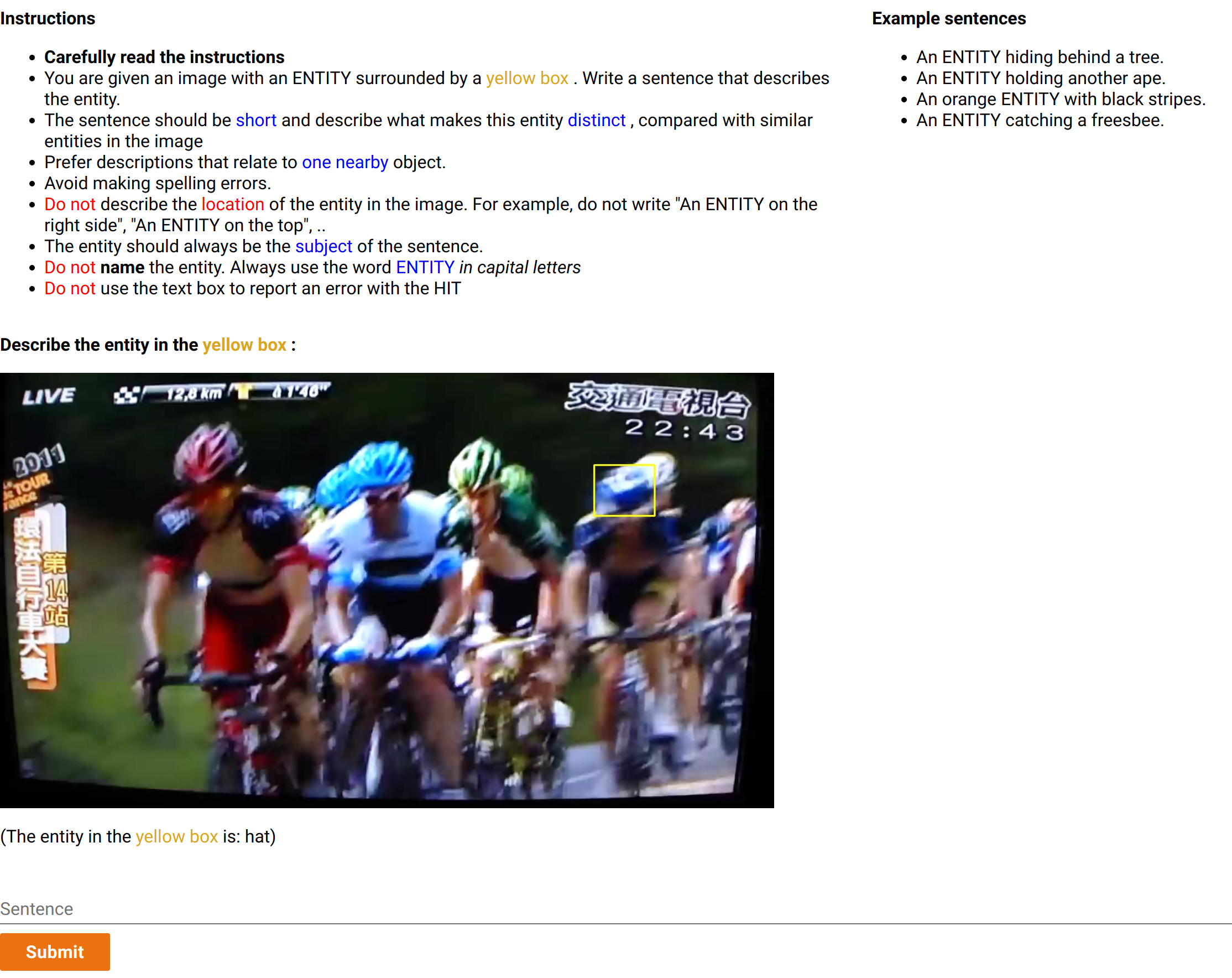} %
    \hspace{-10pt}
    \vspace{-10pt}
    \caption{
Instructions for collecting textual descriptions for object instances in the Youtube-VOS benchmark.
    }
    \label{fig_AMT_describe_entity_segmentation}
    \vspace{-5pt}
\end{figure}

\figref{fig_AMT_describe_entity_segmentation} provides an example of the data collection API for textual annotation of images for the Youtube-VOS benchmark.

\vspace{5pt}\noindent\textbf{Personalized image retrieval:}
We also created an image retrieval variant of YTVOS. We extract a set of images that correspond to the collected captions, where every image in the retrieval set was extracted from a wide box cropped around every instance in each evaluation frame. The box size was set to twice the size of the instance mask \textit{on each axis} (that is, four times the area), to allow it to display some information about the context of the instance.

\end{document}